\documentclass{article}

\PassOptionsToPackage{numbers, compress}{natbib}
\usepackage[preprint]{neurips_2026}
\usepackage[dvipsnames]{xcolor}
\definecolor{airforceblue}{rgb}{0.36, 0.54, 0.66}

\newcommand{\comm}[1]{\iffalse #1 \fi}
\usepackage{colortbl}
\usepackage{graphicx} 
\usepackage{amsmath}
\usepackage{amssymb}
\usepackage{mathtools}
\usepackage{amsthm}
\usepackage{soul}
\usepackage{pifont}
\usepackage{lipsum}
\usepackage{hhline}
\usepackage{booktabs}
\usepackage{subcaption}
\usepackage{multirow}
\usepackage{multicol}
\usepackage{xspace}
\usepackage{float}
\usepackage{algorithm}
\usepackage{algpseudocode}
\algrenewcommand{\algorithmiccomment}[1]{{\footnotesize \hfill // #1}}
\usepackage{bm}
\usepackage{makecell}
\usepackage{adjustbox}
\usepackage[most]{tcolorbox}% For framing the algorithm
\usepackage{enumitem}
\usepackage{wrapfig}
\usepackage{fancyvrb}       % Core package for advanced verbatim environments
\usepackage{fvextra}
\usepackage{xurl}

\newcommand{\xmark}{\ding{55}}%
\definecolor{Gray}{gray}{0.9}
\newcommand{\ours}{E$^2$-TTT\xspace}
\newcommand{\oursmlp}{$\text{E$^2$-TTT}_\text{MLP}$\xspace}
\newcommand{\oursglu}{$\text{E$^2$-TTT}_\text{SwiGLU}$\xspace}
\newcommand{\myparagraph}[1]{\vspace{0pt}\noindent\textbf{#1}}
\definecolor{verylightblue}{RGB}{220,235,245}
\definecolor{lightorange}{RGB}{255, 242, 204}
\definecolor{lightblue}{RGB}{218, 232, 252}
\definecolor{lightgreen}{RGB}{213, 232, 212}

\newcommand{\eg}{\textit{e.g.}}

\theoremstyle{plain}
\newtheorem{theorem}{Theorem}[section]
\newtheorem{proposition}[theorem]{Proposition}

\theoremstyle{definition}

\theoremstyle{remark}

\usepackage[utf8]{inputenc} % allow utf-8 input
\usepackage[T1]{fontenc}    % use 8-bit T1 fonts
\usepackage{url}            % simple URL typesetting
\usepackage{amsfonts}       % blackboard math symbols
\usepackage{nicefrac}       % compact symbols for 1/2, etc.
\usepackage{microtype}      % microtypography

\definecolor{cvprblue}{rgb}{0.21,0.49,0.74}
\usepackage[pagebackref,breaklinks,colorlinks,citecolor=cvprblue]{hyperref}

\title{Rethinking Expressivity and Efficiency in Test-Time Training}

\author{%
  Zeyun Zhong$^{1,3}$ \quad Joya Chen$^2$ \quad Manuel Martin$^3$ \\
  \textbf{Frederik Diederichs$^3$ \quad Juergen Gall$^{4,5}$ \quad Juergen Beyerer$^{1,3}$} \\[5pt]
  $^1$Karlsruhe Institute of Technology (KIT) \\
  $^2$National University of Singapore\\
  $^3$Fraunhofer IOSB \\
  $^4$Lamarr Institute for Machine Learning and Artificial Intelligence \\
  $^5$University of Bonn
}

\begin{document}

\maketitle

\begin{abstract}

Test-Time Training (TTT) enables long-context processing via continuous weight updates during inference, but current methods struggle to balance the hardware efficiency of chunk-wise approximations with the expressivity of per-token update dynamics such as learning rate and momentum. We propose E$^2$-TTT (Expressive and Efficient TTT) to bridge this gap.
Following prior chunk-wise TTT methods, we compute all gradients within a chunk at the chunk-start weights.
Under this setting, we derive a closed-form chunk update that retains these per-token dynamics and exactly reproduces the chunk-end state of the corresponding token-wise recurrence.
This enables fully parallelized chunk-level training while preserving the temporal structure of the update rule that prior chunk-wise methods discard.
We validate E$^2$-TTT by training models up to 1.3B parameters from scratch. 
It performs on par with previous TTT and hybrid attention baselines in language modeling while outperforming them on in-context retrieval. Its advantage is most pronounced in length extrapolation: on the standard ``Needle in a Haystack'' passkey test, it retains over 90\% accuracy at $8\times$ the training context length.
Meanwhile, E$^2$-TTT can match the training throughput of efficient chunk-wise methods, demonstrating that it effectively reconciles expressivity with efficiency.
The code is available at \url{https://github.com/zeyun-zhong/E2-TTT}.

\end{abstract}    
\section{Introduction}
\label{sec:intro}
% Grok4.1,DeepSeek-V3.2,Seed1.8,Qwen3,KimiK2.5

Many applications of large language models (LLMs), including long-horizon agentic tasks~\cite{dong2026longhorizon}, require processing contexts that grow continually over time. Standard Transformers retain this history in a key--value (KV) cache~\cite{kwon2023efficient} and repeatedly attend over it~\cite{vaswani2017attention}, causing memory and computation costs to grow with sequence length. External agent-memory systems mitigate this burden by organizing and selectively retrieving past experiences~\cite{xu2025amem}, but the retrieved memories must still be reintroduced into the model context and therefore do not change the underlying attention and KV-cache scaling. This motivates architectures that incorporate incoming information online into a compact recurrent state without explicitly retaining every preceding token~\cite{gu2024mamba,katharopoulos2020transformers,schlag2021linear,sun2024learning}.

Compact-state models, including linear attention~\cite{katharopoulos2020transformers,yang2024parallelizing}, state-space models (SSMs)~\cite{mamba2,gu2024mamba}, and recurrent components of hybrid architectures~\cite{kimi_linear,qwen3next}, compress history into a fixed-size recurrent state for linear-time processing and parallel training. Gated delta-rule variants enable selective forgetting and targeted overwriting~\cite{yang2024gated}, but finite state capacity still causes memory collisions~\cite{cabannes2026sparse}. From a test-time regression perspective, Gated DeltaNet and test-time training (TTT) share an online-learning interpretation~\cite{wang2025test,yang2024gated}: the former updates a linear associative memory, whereas parametric TTT stores context in the weights of a nonlinear inner learner, such as an MLP, and can use different online optimizers~\cite{sun2024learning}. These \textit{fast weights}~\cite{schlag2021linear} are updated by gradient steps on a self-supervised loss as the sequence is processed. Both approaches remain fixed-size states~\cite{sun2024learning,yang2024gated}, although deep nonlinear fast-weight memories have shown promise in language modeling and long-context retrieval~\cite{behrouz2024titans,behrouz2025atlas} and multimodal generation~\cite{dalal2025one}.

Along the update-granularity spectrum, token-wise TTT lies at the most expressive end~\cite{behrouz2024titans,sun2024learning}: online gradient descent evaluates each token's gradient at fast weights shaped by all preceding updates, yielding a larger effective search space than batch gradient descent~\cite{sun2024learning}, while replacing token-dependent learning-rate, momentum, and decay factors with chunk-shared values loses expressive power~\cite{behrouz2024titans}. This fine-grained state dependence, however, makes the inner loop sequential~\cite{sun2024learning,e2e_ttt}, and small update batches lead to poor parallelism and low hardware utilization~\cite{zhang2025test}. LaCT adopts the opposite operating point, computing the gradient of a summed loss over a large chunk and performing one fast-weight update per chunk, thereby substantially improving GPU utilization~\cite{zhang2025test}. However, LaCT averages per-token momentum factors into a single chunk-level scalar, omitting the position-dependent gradient weighting that token-wise momentum induces within each chunk~\cite{zhang2025test}.

Motivated by this trade-off, we introduce \ours (Expressive and Efficient TTT), a chunk-parallel update that preserves per-token learning rate, momentum, and decay.
Following prior mini-batch and chunk-wise TTT formulations~\cite{sun2024learning,zhang2025test}, we evaluate the gradients of all tokens in a chunk at the fast weights inherited from the previous chunk.
Under this setting, we derive a closed-form chunk update with two scalar kernels that yields exactly the same fast weights and momentum at the end of each chunk as the corresponding token-wise recurrence.
The kernels depend only on per-token scalars and are computed in parallel by log-space cumulative sums, %then applied through two weighted gradient aggregations, 
eliminating the sequential fast-weight recurrence without materializing intermediate matrix states while retaining chunk-level hardware efficiency.
We instantiate \ours in a hybrid architecture in which sliding-window attention captures fine-grained local interactions and the TTT state compresses history across chunks.

We empirically validate \ours by training models of up to 1.3B parameters from scratch on 15B tokens from the
HuggingFace FineWeb-Edu dataset~\cite{penedo2024fineweb}, comparing against strong linear attention and hybrid baselines. Our evaluation spans general language modeling, in-context retrieval, and length extrapolation, where the gains concentrate in the retrieval and extrapolation regimes. 
Notably, our model exhibits strong robustness in length extrapolation, maintaining $>90\%$ accuracy on passkey retrieval in the standard ``Needle in a Haystack'' test at $8\times$ the training context, a regime where prior chunk-based methods collapse. 
Finally, we explore the versatility of our approach through parameter-efficient adaptation for long-context video understanding, achieving performance comparable to full fine-tuning at the 2B-parameter scale on standard multimodal benchmarks. 
%Overall, we demonstrate the benefit of expressive sequential updates while matching the training throughput of standard chunk-wise TTT, offering novel insights into the TTT paradigm.

\section{Background: Test-Time Training}

Conventional recurrent layers compress context into fixed-size vector- or matrix-valued activations. Parametric Test-Time Training (TTT) instead uses the parameters $\mathbf{W}_t$ of an inner learner $f_{\mathbf{W}_t}(\cdot): \mathbb{R}^d \rightarrow \mathbb{R}^d$ as the recurrent state~\cite{sun2024learning}. 
These \textit{fast weights}~\cite{schlag2021linear} are updated by a self-supervised inner loop on each sequence during both training and inference, whereas the \textit{slow weights} are learned by the outer loop and remain fixed at inference. For notational simplicity, we write $\mathbf{W}_t\in\mathbb{R}^{d\times d}$ as a single matrix. For multi-layer learners, $\mathbf{W}_t$ denotes the full collection of fast-weight tensors.

\begin{wrapfigure}{r}{.53\linewidth}
    \vspace{-5mm}
    \centering
    \includegraphics[width=\linewidth]{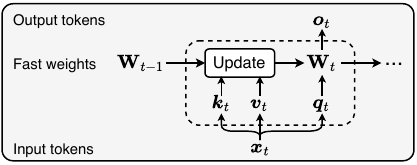}\vspace{-1mm}
    \caption{Token-wise TTT. At each step, the pair $(\boldsymbol{k}_t,\boldsymbol{v}_t)$ defines a self-supervised loss whose gradient step (the ``training'' stage) updates $\mathbf{W}_{t-1}$ to $\mathbf{W}_t$. The query $\boldsymbol{q}_t$ then reads from $\mathbf{W}_t$ to produce $\boldsymbol{o}_t$.}
    \label{fig:ttt_concept}
    \vspace{-4mm}
\end{wrapfigure}
As illustrated in Figure~\ref{fig:ttt_concept}, the general TTT process operates in two stages per step: an update stage and an output stage. 
First, input tokens are projected into query ($\boldsymbol{q}_t$), key ($\boldsymbol{k}_t$), and value ($\boldsymbol{v}_t$) vectors, where $\boldsymbol{q}_t, \boldsymbol{k}_t, \boldsymbol{v}_t \in \mathbb{R}^d$. 
The keys and values are used to update the fast weights via a self-supervised objective, effectively encoding the context into a neural memory with fixed state size. 
Second, the updated fast weights are applied to the query to generate the output token $\boldsymbol{o}_t$. 
Based on how these update and output rules are structured, we categorize current TTT formulations into three groups (see Table~\ref{tab:ttt_comparison_property}).

\myparagraph{Expressive Token-wise Update.} In this standard formulation, both the gradient evaluation and the state update are token-wise. At each time step $t$, the fast weights $\mathbf{W}_t$ are updated by minimizing a self-supervised reconstruction loss $\mathcal{L}$ between the transformed key $f_{\mathbf{W}_{t-1}}(\boldsymbol{k}_t)$ and the value $\boldsymbol{v}_t$. Let $\mathbf{G}_t$ denote the negative gradient, and $\eta_t$ the learning rate:
\begin{align}
    \mathbf{G}_t = - \nabla_{\mathbf{W}} \mathcal{L}\big(f_{\mathbf{W}_{t-1}}(\boldsymbol{k}_t), \boldsymbol{v}_t\big), \qquad \mathbf{W}_t = \mathbf{W}_{t-1} + \eta_t \mathbf{G}_t. \label{eq:ttt_update}
\end{align}
The loss function is commonly Mean Squared Error~\cite{sun2024learning}. 
The updated fast weights $\mathbf{W}_t$ are then immediately used to compute the output vector $\boldsymbol{o}_t$ for the current query $\boldsymbol{q}_t$:
\begin{equation}
    \boldsymbol{o}_t = f_{\mathbf{W}_t}(\boldsymbol{q}_t).
\label{eq:ttt_apply}
\end{equation}

\begin{table}[t]
    \centering
    \caption{Comparison of TTT update schedules. \emph{Gradient evaluated at}: the fast-weight state used to compute the per-token gradients $\mathbf{G}_t$. \emph{State recurrence}: the granularity at which the states $(\mathbf{W}_t, \mathbf{M}_t)$ evolve. \emph{Output computed from}: the state applied to the query $\boldsymbol{q}_t$. \ours matches Chunk-TTT on both efficiency-relevant columns while retaining a token-wise state recurrence.}
    %\vspace{-2mm}
    \label{tab:ttt_comparison_property}
    \setlength\tabcolsep{6pt}
    \begin{tabular}{lcccc}
        \toprule
        Method & \makecell{Gradient \\ Evaluated at} & \makecell{State \\ Recurrence} & \makecell{Output \\ Computed from} & \makecell{Efficient \\ Training} \\
        \midrule
        Token-wise TTT~\cite{sun2024learning} & Previous token & Token-wise & Current token & \xmark \\
        Mini-Batch TTT~\cite{sun2024learning,behrouz2024titans} & Batch start & Token-wise & Current token & \xmark \\
        Chunk-TTT (LaCT)~\cite{zhang2025test} & Chunk start & Chunk-wise & Chunk start & \checkmark \\
        \ours (Ours) & Chunk start & Token-wise & Chunk start & \checkmark \\
        \bottomrule
    \end{tabular}
    \vspace{-4mm}
\end{table}

\noindent\textbf{Token-wise Update with Mini-Batch Gradient.} While the token-wise update offers high expressivity~\cite{sun2024learning}, it cannot parallelize gradient evaluation, since $\mathbf{G}_t$ depends on $\mathbf{W}_{t-1}$. Consequently, recent work~\cite{sun2024learning,behrouz2024titans} adopts a mini-batch approach. The sequence is partitioned into mini-batches indexed by $r$, each of small size $B$ (e.g., 16). We use the notation $(\cdot)_t^{[r]}$ to denote a variable associated with the $t$-th token within the $r$-th batch. 
% Let $\mathbf{W}^{[r-1]} = \mathbf{W}_B^{[r-1]} = \mathbf{W}_0^{[r]}$ denote the state from the previous batch. Inside batch $r$, gradients are approximated using this fixed $\mathbf{W}^{[r-1]}$, which decouples forward pass dependencies:
% \begin{equation}
%     \mathbf{G}_t^{[r]} = - \nabla_{\mathbf{W}} \mathcal{L}\big(f_{\mathbf{W}^{[r-1]}}(\boldsymbol{k}_t^{[r]}), \boldsymbol{v}_t^{[r]}\big), \quad
%     \mathbf{W}_t^{[r]} = \mathbf{W}_{t-1}^{[r]} + \eta_t \mathbf{G}_t^{[r]}, \quad \boldsymbol{o}_t^{[r]} = f_{\mathbf{W}_t^{[r]}}(\boldsymbol{q}_t^{[r]}) 
%     \label{eq:ttt_minibatch_update}
% \end{equation}
Let $\mathbf{W}_B^{[r-1]} = \mathbf{W}_0^{[r]}$ denote the batch-start state inherited from the previous batch. Inside batch $r$, gradients are approximated using this fixed $\mathbf{W}_B^{[r-1]}$, which decouples forward pass dependencies:
\begin{equation}
    \mathbf{G}_t^{[r]} = - \nabla_{\mathbf{W}} \mathcal{L}\big(f_{\mathbf{W}_B^{[r-1]}}(\boldsymbol{k}_t^{[r]}), \boldsymbol{v}_t^{[r]}\big), \quad
    \mathbf{W}_t^{[r]} = \mathbf{W}_{t-1}^{[r]} + \eta_t \mathbf{G}_t^{[r]}, \quad \boldsymbol{o}_t^{[r]} = f_{\mathbf{W}_t^{[r]}}(\boldsymbol{q}_t^{[r]}) 
    \label{eq:ttt_minibatch_update}
\end{equation}
This approximation enables the parallel computation of gradients $\mathbf{G}_t^{[r]}$ for all tokens $t \in \{1, \dots, B\}$. Note that only the state used for gradient evaluation is shared, while the resulting fast weights $\mathbf{W}_t^{[r]}$ and outputs $\boldsymbol{o}_t^{[r]}$ remain token-indexed. Once calculated, the fast weights $\mathbf{W}_t^{[r]}$ can be computed, e.g., via a parallel associative scan~\cite{smith2022simplified}.

\myparagraph{Efficient Chunk-wise Update.} Despite enabling parallel gradient calculation, the small mini-batch size and the token-wise dependency in mini-batch update still result in low hardware utilization (often $<5\%$ FLOPs efficiency~\cite{zhang2025test}). 
LaCT~\cite{zhang2025test} therefore partitions the sequence into large chunks indexed by $r$, each of size $C \gg B$ (e.g., $C=512$). Within the $r$-th chunk, LaCT evaluates all token losses at the chunk-start weights $\mathbf{W}^{[r-1]}$ and leverages the linearity of differentiation to aggregate the optimization objective. Instead of materializing gradients at every token step, it computes the gradient of the accumulated weighted loss, producing a single update direction $\mathbf{G}^{[r]}$ for the entire chunk. When momentum is integrated, LaCT adopts a simplified strategy: averaging the time-dependent momentum factors $\beta_t^{[r]}$ to maintain a chunk-level momentum buffer $\mathbf{M}^{[r]}$:
\begin{gather}
    \mathbf{G}^{[r]} = - \nabla_{\mathbf{W}} \bigg( \sum_{t=1}^C \eta_t^{[r]} \mathcal{L}\big(f_{\mathbf{W}^{[r-1]}}(\boldsymbol{k}_t^{[r]}), \boldsymbol{v}_t^{[r]}\big) \bigg), \label{eq:chunk_gradient} \\
    \mathbf{M}^{[r]} = \bigg( \frac{1}{C} \sum_{t=1}^C \beta_t^{[r]} \bigg) \mathbf{M}^{[r-1]} +  \mathbf{G}^{[r]}, \label{eq:chunk_momentum} \qquad
    \mathbf{W}^{[r]} = \mathbf{W}^{[r-1]} + \mathbf{M}^{[r]}.
\end{gather}
This formulation allows maintaining chunk-level states (i.e., $\mathbf{G}^{[r]}, \mathbf{M}^{[r]}, \mathbf{W}^{[r]}$) instead of token-level states, reducing the number of materialized fast weight states from $C$ to $1$ per chunk. As the new weights $\mathbf{W}^{[r]}$ are only available after processing the entire chunk, the output $\mathbf{O}^{[r]}$ (comprising outputs for all tokens in the chunk) is computed using the weights from the previous chunk. This leaves the fast-weight path blind to intra-chunk history, but renders the chunk forward pass fully parallel:
\begin{equation}
\label{eq:chunkwise_output}
\mathbf{O}^{[r]} = f_{\mathbf{W}^{[r-1]}}(\mathbf{Q}^{[r]}). 
\end{equation}

\section{Expressive and Efficient Test-Time Training}
\label{sec:expressive_chunk_TTT}

Token-wise TTT retains fine-grained optimizer dynamics but is inherently sequential, whereas chunk-wise TTT gains efficiency by collapsing these dynamics into a single update, obscuring the temporal variance of token importance within a chunk.
We introduce \ours (Expressive and Efficient TTT), a closed-form chunk update that preserves the effects of per-token learning rate, momentum, and decay under shared chunk-start gradient evaluation.
It exactly recovers the corresponding chunk-end fast-weight and momentum states while retaining large-chunk parallelism (see Table~\ref{tab:ttt_comparison_property}).
For simplicity, we omit the current-chunk superscript $[r]$ in this section (\eg\ $\mathbf{W}_t=\mathbf{W}^{[r]}_t$). We index the updates within a chunk by $t \in \{1, \dots, C\}$, with $t = 0$ denoting the start state inherited from the previous chunk
%($\mathbf{W}_0 = \mathbf{W}^{[r-1]}$, $\mathbf{M}_0 = \mathbf{M}^{[r-1]}$).
($\mathbf{W}_0 = \mathbf{W}^{[r-1]}_C$, $\mathbf{M}_0 = \mathbf{M}^{[r-1]}_C$).

\subsection{Primal Formulation: Token-wise Dynamics}
\label{sec:primal_formulation}

We first define the token-wise state recurrence that serves as our reference. Unlike the simplified chunk update in Eq.~\ref{eq:chunk_momentum}, it retains per-token momentum and multiplicative decay, motivated by recent work~\citep{behrouz2024titans,yang2024gated}. Let $\mathbf{W}_t, \mathbf{M}_t, \mathbf{G}_t \in \mathbb{R}^{d \times d}$ denote the fast weights, momentum, and negative gradient (update direction), respectively. Let $\gamma_t, \beta_t \in (0,1)$ be the scalar decay and momentum factors, and $\eta_t > 0$ be the learning rate. Our targeted sequential recurrence is defined as:
\begin{equation}
\begin{aligned}
\mathbf{M}_t = \beta_t \, \mathbf{M}_{t-1} + \eta_t \, \mathbf{G}_t, \qquad \mathbf{W}_t = \gamma_t \, \mathbf{W}_{t-1} + \mathbf{M}_t. \label{eq:tokenwise_ttt_decay_momentum}
\end{aligned}
\end{equation}
The specific parameterizations of $\gamma_t$, $\beta_t$, and $\eta_t$ are predicted from the input $\boldsymbol{x}_t$ independently of $(\mathbf{W}, \mathbf{M})$, as described in Section~\ref{sec:parameterization}. As in the mini-batch setting (Eq.~\ref{eq:ttt_minibatch_update}), the gradients $\mathbf{G}_t$ are computed using the fixed weights from the previous chunk 
%($\mathbf{W}_0 = \mathbf{W}^{[r-1]}$)
($\mathbf{W}^{[r-1]}_C$), which decouples them across $t$ and allows them to be computed in parallel before the update step. 
Only the chunk-end states $(\mathbf{W}^{[r]}_C, \mathbf{M}^{[r]}_C)$ are propagated to the next chunk. The intermediate states $(\mathbf{W}^{[r]}_t, \mathbf{M}^{[r]}_t)$ for $t < C$ are not stored.
%are an analytical device and need not be materialized.

\subsection{Closed-Form Parallelization via Scalar Kernels}
\label{sec:closed_form_parallelization}

A naive execution of Eq.~\eqref{eq:tokenwise_ttt_decay_momentum} requires sequential iteration ($t = 1 \to C$). While parallel prefix scans could theoretically compute the states, they typically necessitate 
%materializing 
computing and storing
$\mathbf{W}_t$, $\mathbf{M}_t$, and $\mathbf{G}_t$ at every time step. Since these are dense matrices, storing the full sequence incurs a prohibitive memory cost of $O(C \cdot d^2)$, causing bottlenecks on modern accelerators, especially given the large chunk size $C$.
To enable efficient training without simplifying the dynamics of Eq.~\eqref{eq:tokenwise_ttt_decay_momentum}, we derive a closed-form equivalent that ``jumps'' from the chunk start ($t=0$) to the end ($t=C$) in a single step. Our insight is that while the matrix states $(\mathbf{W}, \mathbf{M})$ are expensive, the decay and momentum factors $(\gamma_t, \beta_t)$ are scalars. By unrolling the recurrence and re-indexing the summation, we can compress the entire history of temporal dynamics into scalar coefficients that isolate the contribution of each gradient $\mathbf{G}_t$ to the final state (see Appendix~\ref{seq:detailed_derivation} for the full derivation):
\begin{equation}
\mathbf{W}_C
= \underbrace{\Big(\prod_{t=1}^{C}\gamma_t\Big)}_{\text{Decay term}} \mathbf{W}_0
 + \underbrace{\sum_{t=1}^{C}\Big(
      \Big(\prod_{i=1}^{t}\beta_i\Big)\mathbf{M}_0 \prod_{q=t+1}^{C}\gamma_q
   \Big)}_{\text{Momentum term}}
 + \underbrace{\sum_{t=1}^{C}\eta_t \mathbf{G}_t
   \sum_{j=t}^{C}\Big(\prod_{i=t+1}^{j}\beta_i \prod_{q=j+1}^{C}\gamma_q\Big)}_{\text{Gradient term}}.
\label{eq:unroll_recurrence}
\end{equation}
The inner summation in the gradient term encapsulates the cumulative impact of the $t$-th token on the final chunk weight. We formalize this by defining efficient suffix products $\widetilde\beta_t, \widetilde\gamma_t$ and a ratio sum $R_t$:
\begin{equation}
\widetilde\beta_t := \prod_{i=t+1}^{C}\beta_i,
\quad
\widetilde\gamma_t := \prod_{i=t+1}^{C}\gamma_i,
\quad
R_t := \sum_{i=t}^{C} \frac{\widetilde\gamma_i}{\widetilde\beta_i},
\label{eq:scalar_aux}
\end{equation}
with the convention $\prod_{i=C+1}^{C}(\cdot) = 1$. These terms depend only on the scalars $\beta_t, \gamma_t$ and can therefore be computed efficiently in parallel (e.g., via log-space cumulative sums) with negligible cost. 
Rewriting the prefix products in Eq.~\eqref{eq:unroll_recurrence} via the identity $\prod_{i=1}^{j}\beta_i = \big(\prod_{i=1}^{C}\beta_i\big) / \widetilde\beta_j$ and unrolling the momentum recurrence in parallel yields the following result.

\begin{proposition}[Exact chunk-end states]
\label{prop:exactness}
Under the recurrence of Eq.~\eqref{eq:tokenwise_ttt_decay_momentum} with $\mathbf{G}_t$ evaluated at the frozen chunk-start $\mathbf{W}_0$ and scalars $\eta_t, \beta_t, \gamma_t$ independent of $(\mathbf{W}, \mathbf{M})$, the chunk-end states satisfy
\begin{align}
\mathbf{M}_C &= \widetilde\beta_0 \, \mathbf{M}_0 + \sum_{t=1}^{C} \mathcal{K}^{M}_t \, \mathbf{G}_t,
& \mathcal{K}^{M}_t &:= \eta_t \, \widetilde\beta_t,
\label{eq:Mc_closed} \\
\mathbf{W}_C &= \widetilde\gamma_0 \, \mathbf{W}_0 + \widetilde\beta_0 R_1 \, \mathbf{M}_0 + \sum_{t=1}^{C} \mathcal{K}^{W}_t \, \mathbf{G}_t,
& \mathcal{K}^{W}_t &:= \eta_t \, \widetilde\beta_t \, R_t.
\label{eq:Wc_closed}
\end{align}
\end{proposition}

\noindent\textbf{Two scalar kernels with distinct temporal weighting.}
Proposition~\ref{prop:exactness} shows that the chunk-end weights $\mathbf{W}_C$ and the chunk-end momentum $\mathbf{M}_C$ aggregate the per-token gradients $\{\mathbf{G}_t\}$ through two distinct scalar kernels. The momentum kernel $\mathcal{K}^{M}_t = \eta_t \widetilde\beta_t$ measures how much of $\mathbf{G}_t$ survives to the chunk end through the momentum decay $\widetilde\beta_t$. The weight kernel $\mathcal{K}^{W}_t = \eta_t \widetilde\beta_t R_t$ additionally weights this by the cumulative weight-decay survival $R_t$, capturing how the gradient propagates through both the momentum buffer and the subsequent weight updates. A bound analysis of $R_t$ and the
resulting kernels $\mathcal{K}^W_t, \mathcal{K}^M_t$ under near-unit $\beta_t, \gamma_t$ is
provided in Appendix~\ref{app:stability}. 
In contrast, LaCT~\citep{zhang2025test} carries only $\eta_t$ in the gradient term (without $\widetilde\beta_t$ or $R_t$) and collapses the momentum factor to a chunk average, removing the position-dependent weighting within a chunk (Eq.~\ref{eq:chunk_momentum}).

\myparagraph{One backward pass for both aggregates.}
Computing the two aggregates $\sum_t \mathcal{K}^{W}_t \mathbf{G}_t$ and $\sum_t \mathcal{K}^{M}_t \mathbf{G}_t$ would naively require two backward passes. We collapse them to one by noting that the kernels differ only in scalar coefficients and share the per-token activation gradients $\boldsymbol{g}_t = -\partial \mathcal{L}_t / \partial f_{\mathbf{W}_0}(\boldsymbol{k}_t)$ (see Appendix~\ref{app:single-backward}). These gradients arise as a $C \times d$ intermediate of LaCT's backward pass, which contracts them into a $d \times d$ weight gradient and then discards them. We instead keep this intermediate and use it for two contractions. The marginal cost over LaCT is therefore a single additional re-weighted contraction. Algorithm~\ref{alg:e2ttt} summarizes the procedure. %as a direct instantiation of Proposition~\ref{prop:exactness}.

The output is computed using the chunk-start weights, $\mathbf{O}^{[r]} = f_{\mathbf{W}_0}(\mathbf{Q}^{[r]})$, consistent with chunk-wise TTT and ensuring a fully parallel forward pass. Algorithm~\ref{alg:e2ttt} retains LaCT's $O(C \cdot d^2)$ per-chunk time complexity: both weighted gradient contractions are $O(C \cdot d^2)$, while the scalar scans add only $O(C)$. We report measured throughput in Appendix~\ref{sec:throughput}.
Cross-chunk equivalence to the original sequential recurrence follows by induction, since $(\mathbf{W}_C, \mathbf{M}_C)$ is precisely the state propagated to the next chunk. We numerically verify this on a 128-chunk trajectory ($C = 512$, 65{,}536 tokens), where the relative $L_2$ deviation from the fully sequential reference is below $2 \times 10^{-6}$ on both $\mathbf{W}_C$ and $\mathbf{M}_C$, within FP32 round-off accumulated over the trajectory.

\begin{algorithm}[t]
\caption{\ours{} chunk update.}
\label{alg:e2ttt}
\begin{algorithmic}[1]
\Require chunk-start state $(\mathbf{W}_0, \mathbf{M}_0)$; tokens $\{(\boldsymbol{q}_t, \boldsymbol{k}_t, \boldsymbol{v}_t)\}_{t=1}^{C}$; scalars $\{\eta_t, \beta_t, \gamma_t\}_{t=1}^{C}$
\State $\boldsymbol{o}_t = f_{\mathbf{W}_0}(\boldsymbol{q}_t), \quad \mathcal{L}_t = \mathcal{L}\big(f_{\mathbf{W}_0}(\boldsymbol{k}_t), \boldsymbol{v}_t\big)$ \Comment{Compute outputs and per-token losses}
\State Compute scalar aggregates $\widetilde\beta_t, \widetilde\gamma_t, R_t$ via log-space cumulative sums; form $\mathcal{K}^{W}_t, \mathcal{K}^{M}_t$
\State $\boldsymbol{g}_t \gets - \partial \mathcal{L}_t / \partial f_{\mathbf{W}_0}(\boldsymbol{k}_t)$ \Comment{Per-token activation gradients (single backward)}

% gradient aggregation
\State $\Delta_W \gets \textsc{Aggregate}\big(\{\mathcal{K}^{W}_t\}, \{\boldsymbol{g}_t\}, \{ \boldsymbol{k}_t\} \big)$ \Comment{$\equiv -\sum_t \mathcal{K}^{W}_t \, \nabla_{\mathbf{W}} \mathcal{L}_t \big|_{\mathbf{W}_0}$}
\State $\Delta_M \gets \textsc{Aggregate}\big(\{\mathcal{K}^{M}_t\}, \{\boldsymbol{g}_t\}, \{\boldsymbol{k}_t\} \big)$ \Comment{$\equiv -\sum_t \mathcal{K}^{M}_t \, \nabla_{\mathbf{W}} \mathcal{L}_t \big|_{\mathbf{W}_0}$; reuses $\{\boldsymbol{g}_t\}$ from line~3}

\State $\mathbf{W}_C \gets \widetilde\gamma_0 \mathbf{W}_0 + \widetilde\beta_0 R_1 \mathbf{M}_0 + \Delta_W$
\State $\mathbf{M}_C \gets \widetilde\beta_0 \mathbf{M}_0 + \Delta_M$
\State \Return outputs $\{\boldsymbol{o}_t\}$, next-chunk state $(\mathbf{W}_C, \mathbf{M}_C)$
\end{algorithmic}
\end{algorithm}

\section{\ours Model Architecture}
\label{sec:architecture}

\begin{wrapfigure}{r}{.5\linewidth}
    \centering
    \vspace{-7mm}
    \includegraphics[width=\linewidth]{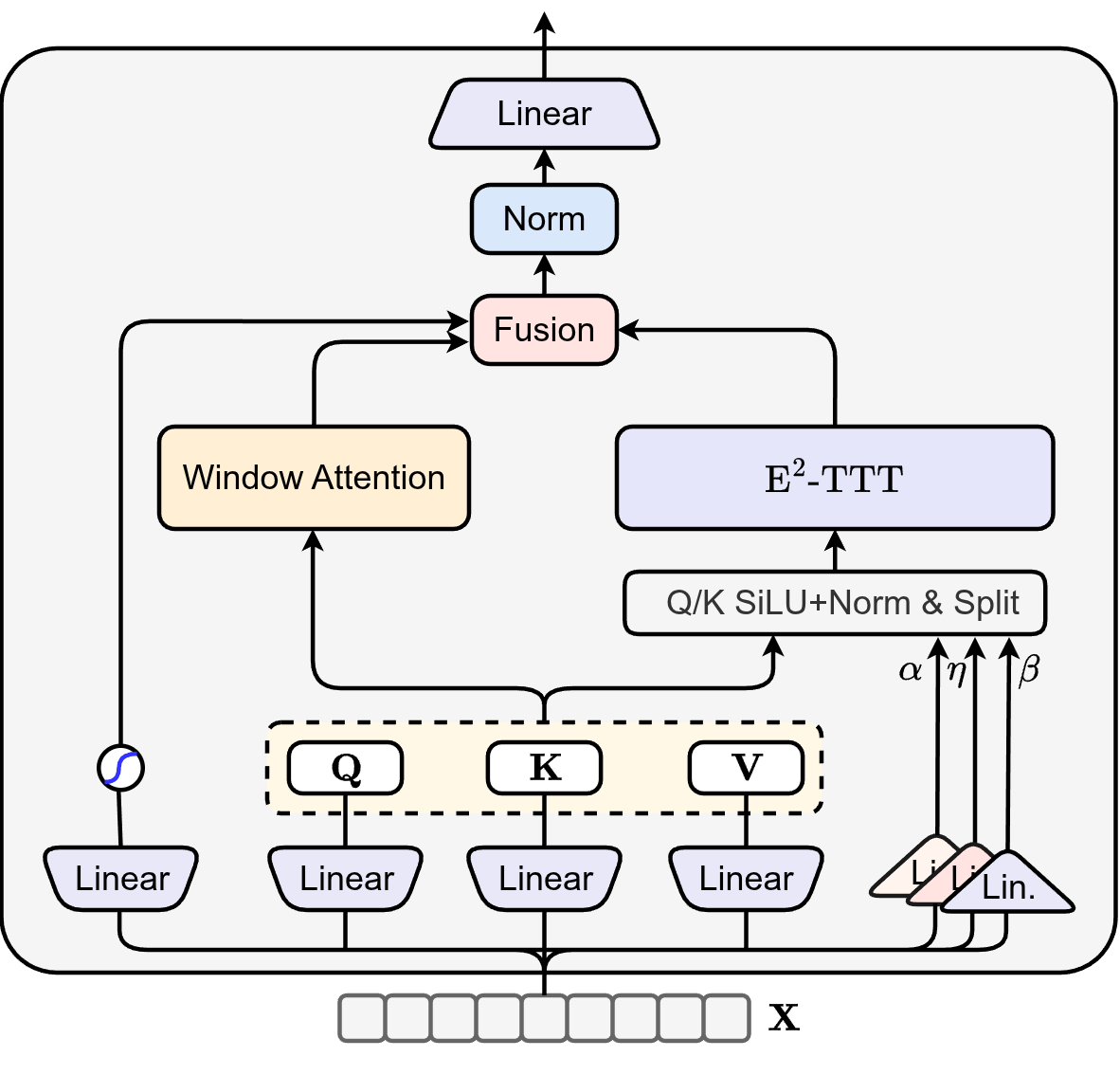}\vspace{-3mm}
    \caption{%\ours Token Mixing Block. The input $\mathbf{X}$ is projected into $\mathbf{Q}, \mathbf{K}, \mathbf{V}$ representations and processed via two parallel branches: Sliding Window Attention for local dependencies and E$^2$-TTT for global context. The TTT update rule is modulated by input-dependent scalars $\alpha, \eta, \beta$, and branch outputs are fused via a data-dependent gate.
    \ours token mixing block. SWA and E$^2$-TTT run in parallel over shared $\mathbf{Q},\mathbf{K},\mathbf{V}$ and are fused by a data-dependent gate. The TTT update is modulated by input-dependent $\alpha,\eta,\beta$.
    }
    \label{fig:attention_block}
    %\vspace{-2mm}
\end{wrapfigure}

The chunk-wise output rule in Eq.~\eqref{eq:chunkwise_output} applies one fast-weight state to every token in a chunk. Consequently, a token cannot influence later tokens in the same chunk through the TTT state. We compensate for this missing within-chunk path with an established hybrid design~\citep{arora2024simple,hua2022transformer,munkhdalai2024leave,zhang2025test}: E$^2$-TTT carries information across chunks, while Sliding Window Attention (SWA) models local causal interactions (Fig.~\ref{fig:attention_block}).

\myparagraph{Overall Architecture.} 
The macro-architecture of \ours follows a Llama-style backbone~\citep{touvron2023llama}, interleaving token-mixing layers with SwiGLU feed-forward blocks. Given $\mathbf{X}\in\mathbb{R}^{L\times d}$, each token-mixing layer projects shared queries $\mathbf{Q}$, keys $\mathbf{K}$, and values $\mathbf{V}$ and sends them to two parallel branches: SWA for local retrieval and E$^2$-TTT for cross-chunk context modeling.
For the TTT branch, we follow prior hybrid and chunk-wise TTT models~\citep{irie2025blending,zhang2025test}, applying SiLU to the queries and keys and then L2-normalizing them for stability. Lightweight linear heads predict the token-dependent learning rate $\eta_t$, momentum factor $\beta_t$, and decay strength $\alpha_t$ from $\mathbf{X}$. The decay factor $\gamma_t$ is then derived from $\eta_t$ and $\alpha_t$, as defined below.
Following HQLT~\citep{irie2025blending}, a token-dependent gate projected from $\mathbf{X}$ interpolates the two branch outputs element-wise. The fused representation is normalized and passed through the output projection.

\myparagraph{Input-Dependent Dynamics.}
\label{sec:parameterization}
We parameterize $\eta_t,\beta_t,\gamma_t$ as functions of $\boldsymbol{x}_t\in\mathbb{R}^d$, allowing the update dynamics to vary across tokens. The corresponding projections are learned end-to-end from the language-modeling objective, without auxiliary supervision. Appendix~\ref{sec:training_details} gives the base scalars, Appendix~\ref{sec:ablation} evaluates their sensitivity, and Appendix~\ref{sec:coeff_analysis} analyzes the learned coefficients.

The learning rate $\eta_t$ modulates the update magnitude. Following TTT~\citep{sun2024learning}, we bound it by a base rate $\eta_{\text{base}}$:
\begin{equation}
\label{eq:eta_parameterization}
\eta_t = \eta_{\text{base}} \cdot \sigma(\text{Linear}_\eta (\boldsymbol{x}_t)) .
\end{equation}
The momentum factor $\beta_t$ controls how strongly past gradients persist. To encourage long-term memory retention, a timescale $\tau$ biases it toward values near $1$, following gated recurrent parameterizations~\citep{yang2023gated,sun2024you}:
\begin{equation}
\beta_t = \sigma(\text{Linear}_\beta(\boldsymbol{x}_t))^{\frac{1}{\tau}}. \label{eq:beta_parameterization}
\end{equation}
The decay factor $\gamma_t$ serves as a forget gate. To prevent excessive regularization when the update step is small, we predict a bounded decay strength $\alpha_t$ and couple it to the learning rate $\eta_t$, yielding the standard multiplicative form of decoupled weight decay~\citep{loshchilov2017decoupled}:
\begin{equation} 
\label{eq:decay_derivation}
\alpha_t = \alpha_{\text{base}} \cdot \sigma(\text{Linear}_\alpha(\boldsymbol{x}_t)), \quad \gamma_t = 1 - \eta_t \alpha_t.
\end{equation}

\myparagraph{Fast-weight Network.}
The formulation supports a general differentiable fast-weight learner $f$. We instantiate it with the two MLPs used in our experiments~\citep{sun2024learning,zhang2025test}: a GELU MLP and a SwiGLU MLP, both with a residual connection and LayerNorm (LN).
The GELU variant uses two fast-weight matrices:
\begin{equation}
f_{\mathbf{W}}(\boldsymbol{x}) = \boldsymbol{x} + \text{LN}\big(\mathbf{W}_2 \, \text{GELU}(\mathbf{W}_1 \boldsymbol{x})\big).
\end{equation}
The SwiGLU variant uses three fast-weight matrices:
\begin{equation}
f_{\mathbf{W}}(\boldsymbol{x})  = \boldsymbol{x} + \text{LN}\big(\mathbf{W}_{3} \, (\text{SiLU}(\mathbf{W}_{1} \boldsymbol{x}) \odot (\mathbf{W}_{2} \boldsymbol{x}))\big).
\end{equation}

\section{Experiments}
\label{sec:experiment}

\begin{table*}[t]
\centering
\caption{General language modeling experiments. We report perplexity ($\downarrow$) and zero-shot accuracy ($\uparrow$) for models with 340M and 1.3B parameters, all trained on 15B tokens. 
\ours attains the lowest perplexity at both scales, while all methods lie
within about two points of one another in average zero-shot accuracy.
Results marked with * are cited from HQLT.}\vspace{-1mm}
\label{tab:commonsense_main}
\setlength\tabcolsep{4pt}
\resizebox{.9\linewidth}{!}{
\begin{tabular}{lc|cc|ccccccc}
\toprule
\textbf{Model} & Attn. & \textbf{Wiki.}  & \textbf{LMB.}  & 
\textbf{LMB.} & \textbf{PIQA} & \textbf{Hella.} & 
\textbf{Wino.} & \textbf{ARC-e} & \textbf{ARC-c} & \textbf{Avg.} \\
& window & ppl $\downarrow$ & ppl $\downarrow$ & {acc} $\uparrow$ & 
{acc} $\uparrow$ & {acc\_n} $\uparrow$ &
{acc} $\uparrow$ & {acc} $\uparrow$ & 
{acc\_n} $\uparrow$ & $\uparrow$ \\
\midrule
\textit{340M params} \\
\quad Transformer++* & 2048 &  26.5 & 34.9 & 33.9 & 67.6 & 41.0 & 53.7 & 60.2 & 29.0 & 47.6 \\
\quad DeltaNet* & -- & 27.6 & 35.0 & 32.8 & 67.1 & 40.8 & 52.6 & 58.5 & 28.8 & 46.8 \\
\quad HQLT* & 512 & 27.0 & 28.0 & 35.9 & 66.3 & 41.3 & 53.2 & 60.1 & 29.0 & 47.6 \\
\quad LaCT & 512 & 27.0 & 28.2 & \textbf{37.1} & 66.8 & 41.6 & 51.4 & 59.0 & 30.0 &  47.7 \\

\quad \oursglu & 512 & 26.4 & 27.8 & 36.6 & \textbf{67.7} & 42.2 & 51.6 & 61.4 & 28.3 & 48.0 \\

\quad \oursmlp & 512 & \textbf{25.5} & \textbf{25.0} & 37.0 & 67.4 & \textbf{43.9} & \textbf{54.9} & \textbf{61.3} & \textbf{30.2} & \textbf{49.1} \\
\midrule

\textit{1.3B params} \\
\quad Transformer++* & 2048 & 19.8 & 17.9 & 42.6 & 71.0 & 50.3 & 55.8 & 65.2 & 33.2 & 53.0  \\
\quad DeltaNet* & -- & 20.6 & 19.9 & 39.3 & 70.1 & 49.5 & 52.5 & 68.5 & 34.2 & 52.3 \\
\quad Mamba2 & -- & 20.5 & 17.0 & 41.1 & 71.0 & 51.5 & 54.5 & 66.5 & 35.0 & 53.3 \\
\quad HQLT & 512 & 20.3 & 16.1 & 42.6 & 71.0 & 51.1 & 55.1 & \textbf{69.2} & 34.0 & 53.8 \\
\quad LaCT & 512 & 20.1 & 16.7 & 42.0 & 70.4 & 51.6 & 55.4 & 65.6 & 33.1 & 53.0 \\
\quad \oursglu & 512 & 19.9 & 15.8 & 43.2 & 71.1 & 51.9 & 54.2 & 67.5 & 34.1 & 53.6  \\
\quad \oursmlp & 512 & \textbf{19.5} & \textbf{15.3} & \textbf{43.7} & \textbf{71.4} & \textbf{52.7} & \textbf{56.1} & 67.6 & \textbf{35.2} & \textbf{54.5} \\
\bottomrule
\end{tabular}}
\vspace{-5mm}
\end{table*}

We evaluate our method across three distinct categories of tasks to verify its versatility and robustness: general language modeling, to validate overall capabilities on standard benchmarks (Sec.~\ref{sec:exp_general_language}); in-context retrieval, to assess the model's precise recall abilities (Sec.~\ref{sec:exp_incontext_retrieval}); and length extrapolation, to test generalization to sequence lengths exceeding the training horizon (Sec.~\ref{sec:exp_extrapolation}). We then isolate the contribution of the update rule itself in a tightly matched ablation (Sec.~\ref{sec:exp_ablation}). 
Finally, we extend our evaluation to long-context video understanding, exploring the potential of our E$^2$-TTT to leverage long-range temporal dependencies in multimodal data (Sec.~\ref{sec:exp_video}).

\myparagraph{Setup.} Following HQLT~\citep{irie2025blending}, we train language models with either 340M or 1.3B trainable parameters (using sequence lengths of 2048 and 2240, respectively) from scratch on 15B tokens of the HuggingFace FineWeb-Edu dataset \citep{penedo2024fineweb}.
These choices mostly follow the configurations of the recent work on linear transformers \citep{yang2023gated,yang2024parallelizing,grazzi2024unlocking}, including the choice of tokenizer (\texttt{fla-hub/transformer-1.3B-100B}).
The subword-unit vocabulary size is 32K for all models.
All models have 24 layers.
The baseline Transformer architecture is from \citet{touvron2023llama} (denoted as Transformer++, following prior convention) and the DeltaNet configuration is from \citet{yang2024parallelizing}. We also compare against two hybrid models: HQLT and LaCT~\citep{zhang2025test}, both of which utilize window attention combined with DeltaNet and chunk-wise TTT, respectively. For all baseline models, we either report the results from the respective papers or train them using official public repositories under identical conditions to \ours for fair comparison. We use a chunk size of 512 for both LaCT and our models. Full architectural, training,
and evaluation details are in Appendix~\ref{sec:training_details}--\ref{sec:evaluation_details}.
Appendix~\ref{app:scaling} additionally reports a feasibility run with a larger training budget.

\subsection{General Language Modeling Capabilities}
\label{sec:exp_general_language}

In Table~\ref{tab:commonsense_main}, we present the language modeling perplexity and zero-shot accuracy on commonsense reasoning benchmarks for models with 340M and 1.3B parameters. 
\ours obtains the lowest perplexity at both scales, with the margin most visible on LAMBADA (15.3 vs.\ 16.1 for the strongest baseline at 1.3B, and 25.0 vs.\ 28.0 at 340M).
Gains in zero-shot accuracy are smaller: \oursmlp reaches 54.5\% average at 1.3B against 53.8\% for the strongest baseline, and leads on five of six tasks while trailing HQLT and DeltaNet on ARC-e.

\subsection{Evaluating In-Context Retrieval Abilities}
\label{sec:exp_incontext_retrieval}

\begin{wraptable}{r}{.52\linewidth}
\centering
\vspace{-5mm}
\caption{Accuracy on real-world retrieval tasks with input truncated to 2K tokens. $^*$: from HQLT.\comm{Results with * are taken from HQLT.}}
\label{tab:incontext_retrieval_main}
\vspace{-2mm}
\setlength\tabcolsep{4pt}
\resizebox{\linewidth}{!}{
\begin{tabular}{lc|cccc}
\toprule
\textbf{Model} & Attn. & \textbf{SWDE} & \textbf{SQuAD} & \textbf{FDA} & \textbf{Avg.} \\
& window & acc $\uparrow$ & acc $\uparrow$ & acc $\uparrow$ & \\
\midrule
\textit{340M params} \\
\quad Transformer++* & 2048 & \textbf{44.9} & 36.9 & \textbf{52.3} & \textbf{44.7} \\
\quad Transformer++* & 1024 & 30.4 & 25.5 & 31.2 & 29.0 \\
\quad DeltaNet*      & --   & 18.5 & 25.2 & 8.6  & 17.4 \\
\quad HQLT* & 512 & 22.9 & 35.7 & 17.3 & 25.3 \\
\quad LaCT & 512 & 25.0 & 39.0 & 22.2 & 28.7 \\
\quad \oursglu & 512 & \underline{35.4} & \underline{39.6} & \underline{26.1} & \underline{33.7}  \\
\quad \oursmlp & 512 & 23.6 & \textbf{40.1} & 26.0 & 29.9 \\
\midrule
\textit{1.3B params} \\
\quad Transformer++* & 2048 & \textbf{53.7} & 41.5 & \textbf{64.7} & \textbf{53.3}\\
\quad DeltaNet* & -- & 32.9 & 29.9 & 23.6 & 28.8 \\
\quad Mamba2 & -- & 33.8 & 37.3 & 26.1 & 32.4 \\
\quad HQLT & 512 & 32.1 & 43.4  & 31.0 & 35.5 \\
\quad LaCT & 512 & 39.2 & \underline{44.9} & 26.0 & 36.7 \\
\quad \oursglu & 512 & \underline{47.3} & 44.4 & \underline{39.0} & \underline{43.6} \\
\quad \oursmlp & 512 & 37.2 & \textbf{45.6} & 28.5 & 37.1  \\
\bottomrule
\end{tabular}}
\vspace{-4mm}
\end{wraptable}
In this section, we evaluate the performance of \ours on recall-intensive tasks.
Following prior work \citep{yang2024parallelizing,arora2023language,irie2025blending},
we focus on three tasks: FDA \citep{arora2023language}, SWDE \citep{lockard2019openceres}, and SQuAD \citep{rajpurkar2018know}. Table~\ref{tab:incontext_retrieval_main} summarizes the results for the 340M and 1.3B parameter scales.
Consistent with~\citet{irie2025blending}, full attention models (Transformer++) still hold an advantage (53.3\% average) due to their explicit history access. However, \ours substantially narrows this gap: our 1.3B SwiGLU model (43.6\%) surpasses HQLT (35.5\%) and LaCT (36.7\%) on average, demonstrating the benefit of our expressive chunk-wise update rule. Our SwiGLU variant also outperforms the MLP variant by a large margin (43.6\% vs.\ 37.1\% average), indicating that gating mechanisms are beneficial for precise information extraction.

\begin{figure*}[t]
    \centering
    \includegraphics[width=.95\linewidth]{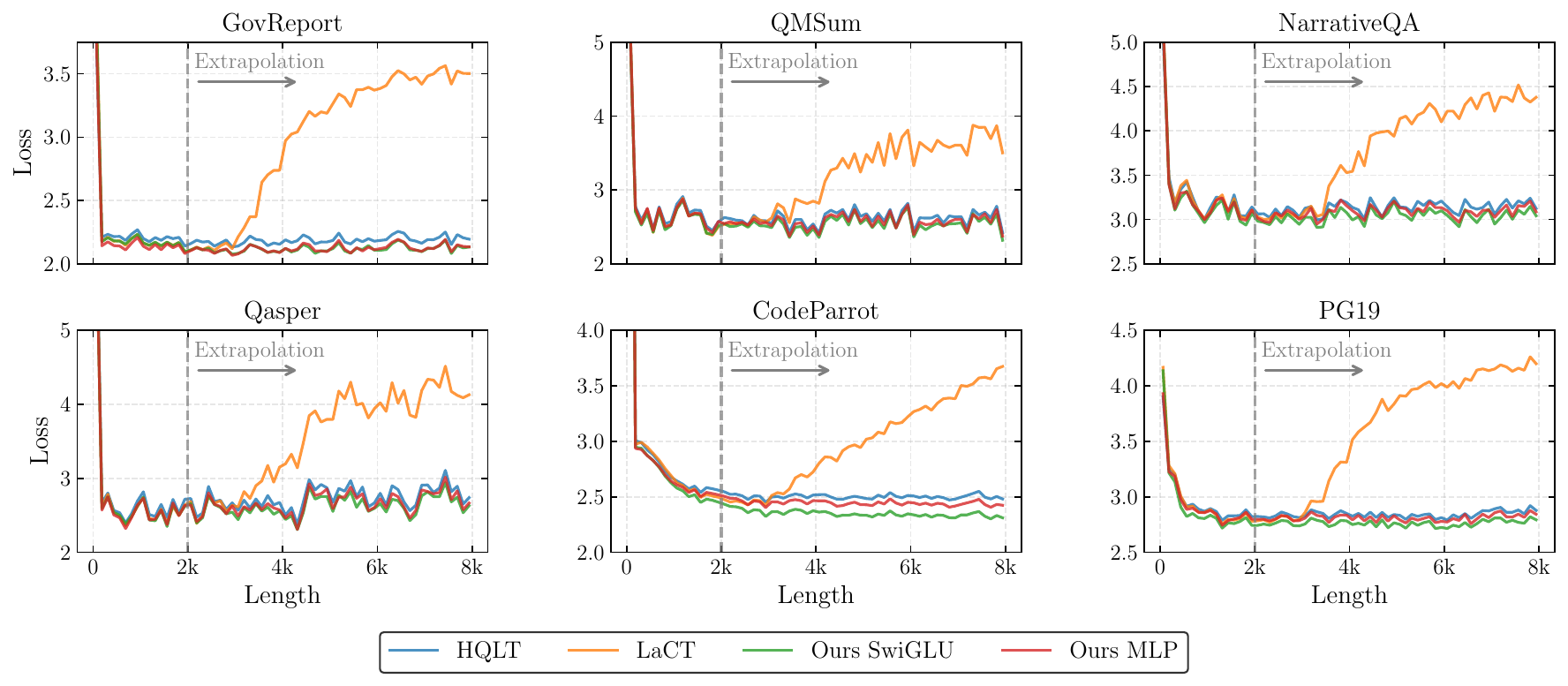}\vspace{-1mm}
    \caption{
    Length extrapolation of 1.3B models. We visualize the per-token loss metric~\citep{xiong2024effective} on six long-context benchmarks, where \ours maintains stability while the LaCT baseline diverges.
    }%\vspace{-4mm}
    \label{fig:loss_curve}
\end{figure*}

\subsection{Evaluating Length Extrapolation Capabilities}
\label{sec:exp_extrapolation}

A critical advantage of our \ours is its potential to generalize to sequence lengths significantly exceeding the training horizon. We evaluate this capability across three dimensions: language modeling stability, synthetic long-context retrieval, and real-world long-context benchmarks.

\myparagraph{Language Modeling Extrapolation.} We first assess whether the models can maintain stable language modeling performance when processing sequences longer than their training context (2K tokens). Figure~\ref{fig:loss_curve} visualizes the per-token loss metric~\citep{xiong2024effective} on six long-context datasets. As shown in Fig.~\ref{fig:loss_curve}, LaCT struggles with length extrapolation, with loss exploding immediately past the training boundary on all six datasets. HQLT maintains stability but shows higher validation loss (e.g., on CodeParrot). In contrast, both \ours variants demonstrate lower and stable loss curves throughout the extrapolated region. This confirms that our expressive chunk-wise updates robustly generalize to unseen lengths.

\begin{wrapfigure}{r}{.65\linewidth}
    \centering
    \vspace{-5mm}
    \includegraphics[width=\linewidth]{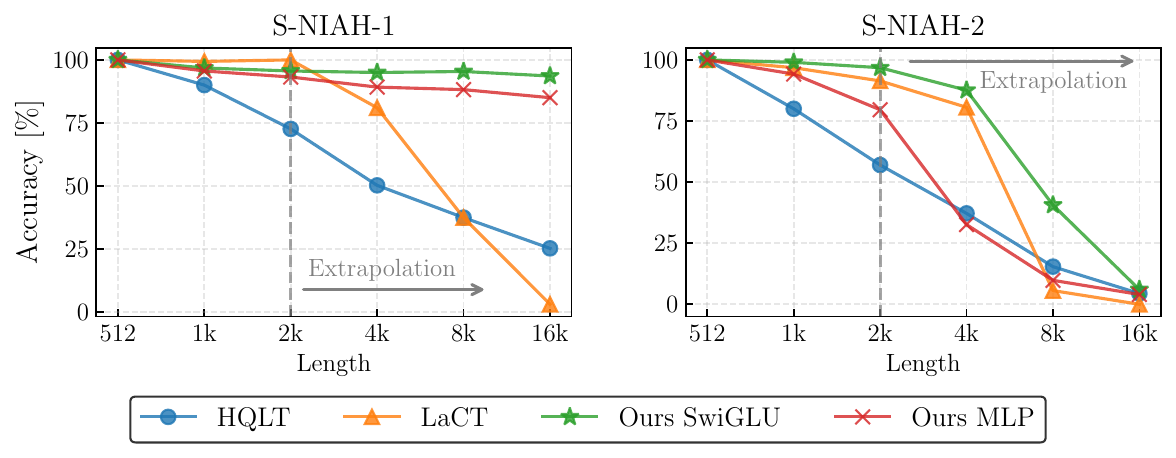}\vspace{-1mm}
    \caption{Zero-shot accuracy on S-NIAH-1 and S-NIAH-2 for 1.3B models. All methods use sliding window attention with a 512-token window. \ours maintains robust retrieval well beyond the training context, while LaCT collapses past 4K and HQLT degrades steadily across both tasks.}
    \label{fig:single_niddle}
    \vspace{-2mm}
\end{wrapfigure}
\myparagraph{Long-range Associative Recall.} Stability alone does not guarantee information retention. To test effective capacity, we employ the ``Single-Needle in a Haystack'' (S-NIAH) benchmark~\citep{hsieh2024ruler}, extending the test sequence up to 16K tokens (8$\times$ the training horizon). Figure~\ref{fig:single_niddle} reveals a stark contrast in extrapolation behavior. On S-NIAH-1, both \ours variants retain at least 85\% accuracy at 16K, while LaCT collapses to near-zero and HQLT degrades to 25\%. On S-NIAH-2, \oursglu retains a clear lead
through 16K, achieving 40.6\% at 8K versus 15.4\% (HQLT) and 5.6\% (LaCT). The SwiGLU variant consistently outperforms the MLP variant in the extrapolation regime, indicating the benefit of a gated fast-weight network for handling complex retrieval patterns. A more detailed numerical comparison, including additional baselines, 
is provided in Appendix~\ref{app:s-niah-full}.

\begin{table*}[t]
\centering
\caption{Evaluation on Longbench. Accuracy on 14 tasks from LongBench: Narrative QA, QasperQA, MultiField QA, HotpotQA, 2WikiMulti QA, Musique, GovReport, QMSum, MultiNews, TRec, Trivia QA, SamSum, LCC, and RepoBench-P by order.}
\label{tab:longbench}
\setlength\tabcolsep{4pt}
\resizebox{\linewidth}{!}{
\begin{tabular}{l|ccc|ccc|ccc|ccc|cc|c}
\toprule
\multirow{2}{*}{Model (1.3B)} & \multicolumn{3}{c|}{Single-Doc QA} & \multicolumn{3}{c|}{Multi-Doc QA} & \multicolumn{3}{c|}{Summarization} & \multicolumn{3}{c|}{Few-shot} & \multicolumn{2}{c|}{Code} & \multirow{2}{*}{Avg.} \\
 & NQA & QQA & MFQ & HQA & 2WM & Mus & GvR & QMS & MNs & TRC & TQA & SSM & LCC & RBP &  \\ 
\midrule
Mamba2 & 2.0 & 4.6 & 11.4 & 3.5 & 8.9 & 2.8 & 4.7 & 16.6 & 7.7 & 9.5 & 16.6 & 13.0 & 22.3 & 20.0 & 10.3 \\
HQLT & 0.8 & 1.0 & 9.3 & 3.3 & 7.9 & 2.0 & 6.7 & 14.5 & 12.4 & 25.8 & 19.9 & \textbf{23.1} & 19.8 & 22.8 & 12.1 \\
LaCT & 1.1 & 4.5 & 8.2 & 3.5 & 4.7 & 2.4 & 9.4 & 5.1 & 11.5 & 10.0 & 5.6 & 7.6 & 19.4 & 14.5 & 7.7 \\
\oursglu & \textbf{2.8} & \textbf{4.9} & \textbf{14.2} & \textbf{5.0} & 9.0 & \textbf{3.8} & \textbf{10.3} & \textbf{16.7} & 12.1 & \textbf{34.0} & 16.2 & 20.1 & \textbf{24.4} & 23.2 & \textbf{14.1} \\
\oursmlp & 1.7 & 4.4 & 12.4 & 4.4 & \textbf{10.4} & 3.1 & 8.8 & 16.6 & \textbf{13.6} & 26.5 & \textbf{20.3} & 19.3 & 19.5 & \textbf{24.6} & 13.3 \\
\bottomrule
\end{tabular}}
\vspace{-4mm}
\end{table*}

\myparagraph{Real-world Long Context Understanding.} Finally, we evaluate performance on LongBench~\citep{bai2024longbench}, a comprehensive suite of 14 real-world long-context tasks. As detailed in Table~\ref{tab:longbench}, our method translates its length extrapolation advantages into practical gains. \oursglu achieves the highest average score (14.1\%), outperforming HQLT (12.1\%) and the SSM-based Mamba2 (10.3\%), while nearly doubling the performance of LaCT (7.7\%). The improvements are systematic across task categories: \oursglu outperforms Mamba2 on 13 of 14 tasks, HQLT on 11 of 14, and LaCT on all 14. The MLP variant similarly outperforms all baselines (13.3\% average), confirming that the gains stem from our update rule rather than a specific gating choice. This consistency across single-document QA, multi-document QA, summarization, few-shot, and code tasks demonstrates that our expressive chunk-wise update rule enables \ours to leverage extended context more effectively than baselines spanning multiple sub-quadratic paradigms (SSM, linear attention, and TTT).

\subsection{Ablation Study}
\label{sec:exp_ablation}

\begin{wraptable}{r}{.55\linewidth}
\centering
\vspace{-4mm}
\caption{Controlled comparison of update rules (1.3B). The last three rows share
our framework and differ only in how the inner loop is aggregated. Training context
is 2K tokens. Columns to the right of 2K are extrapolation.}
\label{tab:matched_lact}
\setlength\tabcolsep{4pt}
\resizebox{\linewidth}{!}{
\begin{tabular}{l|cccc|ccc}
\toprule
\multirow{2}{*}{Method} & \multicolumn{4}{c|}{S-NIAH-1} & \multicolumn{3}{c}{S-NIAH-2} \\
 & 2K & 4K & 8K & 16K & 2K & 4K & 8K \\
\midrule
SWA only              & 26.8 & -- & 6.8 & 2.8 & -- & -- & -- \\
\midrule
LaCT     & \textbf{100.0} & 81.0 & 37.4 & 3.0 & 91.4 & 80.6 & 5.6 \\
LaCT-Matched          & 99.8 & \textbf{99.2} & 51.6 & 0.0 & \textbf{98.6} & 65.8 & 4.8 \\
Chunk-averaged        & 25.2 & 10.8 & 6.8 & 3.0 & 30.4 & 18.2 & 7.2 \\
\oursglu              & 95.6 & 95.0 & \textbf{95.4} & \textbf{93.6} & 96.8 & \textbf{87.6} & \textbf{40.6} \\
\bottomrule
\end{tabular}}
\vspace{-3mm}
\end{wraptable}

To isolate the update rule from the surrounding architecture, we train two additional 1.3B models
inside our own framework. Apart from the published LaCT configuration and the SWA-only
reference, all rows of Table~\ref{tab:matched_lact} share the same architecture, including
SWA and the fusion gate, and differ only in how the inner loop is aggregated. LaCT-Matched ports LaCT's update rule into
our framework. Chunk-averaged is our own model with the
per-token momentum and decay factors collapsed to chunk-level scalars. 
LaCT-Matched improves over its default configuration on S-NIAH-1 within and
near the training context (81.0$\to$99.2 at 4K) and is
slightly ahead of our own model at 2K, while on S-NIAH-2 it is mixed
(98.6 vs.\ 91.4 at 2K, 65.8 vs.\ 80.6 at 4K).
Beyond the training context, however, it still collapses to 0.0 at
16K. Chunk-averaged isolates the effect of the per-token scalars.
Collapsing them drops S-NIAH-1 at 8K from 95.4 to 6.8, the level of the SWA
baseline alone, and the degradation already
begins inside the training context (95.6$\to$25.2 at 2K). The collapsed model nonetheless trains normally and is even marginally ahead in-domain
(15.6 vs.\ 15.8 LAMBADA perplexity, 53.8\% vs.\ 53.6\% zero-shot average).
We therefore attribute the long-context retrieval gains, both within and beyond the
training context, to our expressive chunk-wise update rule.
Ablations of the remaining components, including inner-loop optimizer terms, chunk granularity, and
sensitivity to the base scalars, are reported in
Appendix~\ref{sec:ablation}.

\subsection{Multimodal Evaluation}
\label{sec:exp_video}

\begin{wraptable}{r}{.6\linewidth}
    \centering
    \vspace{-5mm}
    \caption{Video understanding benchmarks. 
    Integrating \ours into Qwen3VL-2B-Instruct and training only the TTT parameters matches full fine-tuning of the backbone on the same training data.}
    %\vspace{-1mm}
    \label{tab:video_understanding_2B}
    \setlength\tabcolsep{4pt}
    \resizebox{\linewidth}{!}{
    \begin{tabular}{lccccccc}
    \toprule
    \multirow{2}{*}{Model} &  \multicolumn{4}{c}{VideoMMMU} & \multicolumn{3}{c}{LongVideoBench}\\
    \cmidrule(lr){2-5} \cmidrule(lr){6-8}
    & Perc. & Comp. & Adap. & Avg. & Perc. & Rel. & Avg. \\
    \midrule
    Qwen3VL & --& -- & -- & 41.9 & 63.1 & 51.5 & 56.9 \\
    Qwen3VL (SFT) & 60.3 & \textbf{37.3} & 33.0 & 43.5   & 63.4 & 54.0 & 58.3 \\
    Qwen3VL+\ours  & \textbf{60.7} & \textbf{37.3} & \textbf{33.7} & \textbf{43.9}  & \textbf{63.7} & \textbf{54.9} & \textbf{59.0} \\
    \bottomrule
    \end{tabular}}
    \vspace{-2mm}
\end{wraptable}
To assess generalization beyond language modeling, we integrate \ours into Qwen3VL-2B-Instruct~\citep{Qwen3-VL} as a parallel branch fused with self-attention layers. We freeze the pre-trained backbone and train only the TTT parameters on a 43K-sample subset of LLaVA-Video-178K~\citep{zhang2024video}. 
Qwen3VL+\ours improves over the frozen base model (43.9 vs.\ 41.9 on VideoMMMU,
59.0 vs.\ 56.9 on LongVideoBench) and matches a fully fine-tuned baseline trained on the
same data (43.5 and 58.3, respectively).
Additional training and evaluation details are provided in Appendix~\ref{sec:training_details} and \ref{sec:evaluation_details}.

\section{Related Work}
\label{sec:related_work}

Test-Time Training (TTT)~\cite{sun2024learning} is an emerging paradigm in sequence modeling that redefines the recurrent states in RNNs as learnable weights of an online-adapted non-linear neural network. These weights, often referred to as fast weights~\cite{schlag2021linear}, are continuously updated to learn in-context. Existing methods typically employ a self-supervised loss that encourages these fast weights to memorize key-value associations from in-context tokens, effectively treating inference as an online learning problem using variants of gradient descent. While TTT~\cite{sun2024learning,wang2025test} has motivated a range of new recurrent architectures~\cite{behrouz2024titans,karami2025lattice,behrouz2025atlas,e2e_ttt}, its token-wise updates are inherently sequential, leading to poor hardware utilization and limiting scalability. Recent efforts like LaCT~\cite{zhang2025test} have attempted to mitigate this by shifting from token-wise to chunk-wise updates, sacrificing intra-chunk dependency modeling to improve throughput. However, these methods often resort to simplified aggregation rules (e.g., flat averaging) that degrade the optimization fidelity. Our work builds upon this foundation but distinguishes itself by introducing \ours (Expressive and Efficient TTT), a formulation that reconciles the hardware efficiency of chunking with the mathematical precision of complex, token-wise update dynamics.
Combining global recurrence with local attention is a proven strategy for efficient long-context modeling~\cite{lieber2024jamba,ren2024samba,de2024griffin}. Following recent work~\cite{irie2025blending,zhang2025test}, we adopt a \textit{within-layer} hybrid architecture that shares the key, value, and query vectors between local attention and global recurrence. This design pattern has been successfully deployed in various forms~\cite{hua2022transformer,munkhdalai2024leave,arora2023language}. However, prior instantiations typically rely on standard linear attention~\cite{katharopoulos2020transformers,schlag2021linear,yang2024parallelizing,irie2025blending} or simplified TTT update rules~\cite{zhang2025test}. Our work retains this established architectural template but replaces the recurrence module with our \ours, empirically showing superior results.
\section{Conclusion}

We presented \ours (Expressive and Efficient TTT), a method that reconciles the hardware efficiency of chunk-wise processing with the per-token expressivity of the update rule in Test-Time Training. 
By deriving two closed-form scalar kernels, we enabled the parallel execution of per-token recurrent dynamics without the chunk-level approximations of prior chunk-wise methods. 
Empirically, \ours is on par with strong sub-quadratic baselines in language modeling and outperforms them on retrieval while demonstrating superior length extrapolation, sustaining $>90\%$ passkey retrieval accuracy on ``Needle in a Haystack'' at $8\times$ the training context length, where baseline methods collapse. These results suggest that preserving precise temporal dynamics is essential for long-context generalization, positioning \ours as a promising direction for scalable Test-Time Training.

\section*{Acknowledgements}
This work was supported by the JuBot project funded by the Carl-Zeiss-Foundation.
The authors acknowledge support by the state of Baden-W\"urttemberg through bwHPC.
We acknowledge EuroHPC Joint Undertaking for awarding us access to MareNostrum5 at BSC, Spain.
Experiments were performed on the HoreKa supercomputer, funded by the Ministry of Science, Research and the Arts Baden-W\"urttemberg and by the Federal Ministry of Education and Research, and on MareNostrum5.
Juergen Gall has been supported by the Federal Ministry of Research, Technology and Space (BMFTR) under grant no.~16IS22094A
WEST-AI and the ERC Consolidator Grant FORHUE (101044724).

{
    \small
    \bibliographystyle{ieeenat_fullname}
    \bibliography{main}
}

\newpage
\newpage
\appendix

\section{Detailed Derivation of the Parallel Kernel}
\label{seq:detailed_derivation}

In this section, we derive the closed-form expansion of the TTT update rule within a single chunk of length $C$. This formulation allows us to ``jump'' directly from the start of the chunk ($t = 0$) to the end ($t = C$) without materializing intermediate states. The chunk index $r$ is omitted for simplicity. 

Recall that we consider general per-token updates with decoupled weight decay and momentum. Let the fast weights $\mathbf{W}_t\in\mathbb{R}^{d\times d}$, momentum $\mathbf{M}_t\in\mathbb{R}^{d\times d}$, and update direction (negative gradient) $\mathbf{G}_t\in\mathbb{R}^{d\times d}$ be matrix-valued. Let $\gamma_t, \beta_t \in (0,1)$ be the scalar decay and momentum factors, and $\eta_t > 0$ be the learning rate. The recurrence is:
\begin{equation}
    \mathbf{M}_t = \beta_t \, \mathbf{M}_{t-1} + \eta_t \, \mathbf{G}_t, \qquad
    \mathbf{W}_t = \gamma_t \, \mathbf{W}_{t-1} + \mathbf{M}_t. \label{eq:tokenwise_ttt_decay_momentum_appendix}
\end{equation}

\subsection{Unrolling Momentum and Weights}

\myparagraph{Unrolling the momentum.} By induction, the momentum $\mathbf{M}_t$ at step $t$ is the sum of the decayed initial momentum at the chunk start and the history of gradients:
\begin{equation}
\label{eq:m-unroll-matrix-scalar}
\mathbf{M}_t
= \Big(\prod_{i=1}^{t}\beta_i\Big) \mathbf{M}_0
  + \sum_{j=1}^{t}\Big(\eta_j\, \mathbf{G}_j \prod_{i=j+1}^{t}\beta_i\Big).
\end{equation}
\myparagraph{Unrolling the Weights.} Similarly, unrolling the weight update (Eq.~\ref{eq:tokenwise_ttt_decay_momentum_appendix}) yields:
\begin{align}
\mathbf{W}_t = \Big(\prod_{i=1}^{t}\gamma_i\Big) \mathbf{W}_0
   + \sum_{j=1}^{t}\Big(\mathbf{M}_j \prod_{i=j+1}^{t}\gamma_i\Big).
\label{eq:w-unroll-stage1-matrix-scalar}
\end{align}
Insert \eqref{eq:m-unroll-matrix-scalar} into \eqref{eq:w-unroll-stage1-matrix-scalar} and separate the
\(\mathbf{M}_0\) and gradient parts:
\begin{equation}
\mathbf{W}_t
= \underbrace{\Big(\prod_{i=1}^{t}\gamma_i\Big) \mathbf{W}_0}_{\text{Decay term}} 
 + \underbrace{\sum_{j=1}^{t}\Big(
      \Big(\prod_{i=1}^{j}\beta_i\Big)\mathbf{M}_0 \prod_{q=j+1}^{t}\gamma_q
   \Big)}_{\text{Momentum term}} 
 + \underbrace{\sum_{j=1}^{t}\sum_{k=1}^{j}
    \Big(\eta_k \mathbf{G}_k \prod_{i=k+1}^{j}\beta_i \prod_{q=j+1}^{t}\gamma_q\Big)}_{\text{Gradient term}}.
\label{eq:w-unroll-stage2-matrix-scalar}
\end{equation}
\myparagraph{Re-indexing the gradient contribution.}
Swap the order of summation in the last term of \eqref{eq:w-unroll-stage2-matrix-scalar} by
fixing the gradient index $k$ and summing over \(j=k,\dots,t\):
\begin{equation}
\sum_{j=1}^{t}\sum_{k=1}^{j}
\Big( \eta_k \mathbf{G}_k \prod_{i=k+1}^{j}\beta_i \prod_{q=j+1}^{t}\gamma_q \Big) = \sum_{k=1}^{t} \eta_k \mathbf{G}_k \sum_{j=k}^{t}\Big(\prod_{i=k+1}^{j}\beta_i \prod_{q=j+1}^{t}\gamma_q\Big).
\label{eq:swap-matrix-scalar}
\end{equation}
This converts Eq.~\ref{eq:w-unroll-stage2-matrix-scalar} to:
\begin{equation}
\mathbf{W}_t
= \underbrace{\Big(\prod_{i=1}^{t}\gamma_i\Big) \mathbf{W}_0}_{\text{Decay term}} 
 + \underbrace{\sum_{j=1}^{t}\Big(
      \Big(\prod_{i=1}^{j}\beta_i\Big)\mathbf{M}_0 \prod_{q=j+1}^{t}\gamma_q
   \Big)}_{\text{Momentum term}} 
   + \underbrace{\sum_{k=1}^{t}\eta_k \mathbf{G}_k
   \sum_{j=k}^{t}\Big(\prod_{i=k+1}^{j}\beta_i \prod_{q=j+1}^{t}\gamma_q\Big)}_{\text{Gradient term}}.
\label{eq:w-unroll-stage2-matrix-scalar_final}
\end{equation}

\subsection{Closed form at the final step per chunk (\(t=C\))}

To efficiently compute the final state $\mathbf{W}_C$, we define three auxiliary scalar sequences. Let the suffix products for decay $\gamma$ and momentum $\beta$ be:
\begin{equation}
\label{eq:suffix-products-scalar}
\tilde\beta_t := \prod_{i=t+1}^{C}\beta_i,
\qquad
\tilde\gamma_t := \prod_{i=t+1}^{C}\gamma_i,
\end{equation}
(with the empty product convention $\prod_{C+1}^{C} (\cdot) = 1$).
We further define the cumulative ratio sum $R_t$:
\begin{equation}
\label{eq:R-def-scalar}
R_t := \sum_{i=t}^{C} \frac{\tilde\gamma_i}{\tilde\beta_i}.
\end{equation}
Since these terms depend only on the scalars $\beta, \gamma$, they can be pre-computed efficiently in parallel (e.g., via log-space cumulative sums).

\myparagraph{Initial Weight Carry ($\mathrm{carry}_\mathrm{w}$).} Directly from definition:
\begin{equation}
\label{eq:carry-u-scalar}
\mathrm{carry}_\mathrm{w} := \prod_{i=1}^{C}\gamma_i = \tilde\gamma_0
\end{equation}
\myparagraph{Initial Momentum Carry ($\mathrm{carry}_\mathrm{m}$).} We rewrite the prefix product $\prod_{i=1}^{j}\beta_i$ as the total product divided by the suffix product:
\begin{equation}
    \prod_{i=1}^{j}\beta_i = \big(\prod_{i=1}^{C}\beta_i\big)\big/ \tilde\beta_j.
\end{equation}
Substituting this into the $\mathbf{M}_0$ term in Eq.~\ref{eq:w-unroll-stage2-matrix-scalar_final}:
\begin{equation}
\sum_{j=1}^{C}\Big(\big(\prod_{i=1}^{j}\beta_i\big)\, \mathbf{M}_0\, \tilde\gamma_j\Big)
= \Big(\prod_{i=1}^{C}\beta_i\Big)
  \underbrace{\sum_{j=1}^{C}\frac{\tilde\gamma_j}{\tilde\beta_j}}_{=\,R_1}
  \, \mathbf{M}_0,
\label{eq:carry-m-scalar}
\end{equation}
with
\begin{equation}
    \mathrm{carry}_\mathrm{m} := \tilde\beta_0 R_1.
\end{equation}
\myparagraph{Gradient contributions.}
From \eqref{eq:swap-matrix-scalar} at \(t=C\), each \(\mathbf{G}_k\) is weighted by the scalar
\begin{equation}
\sum_{j=k}^{C}
\Big(\prod_{i=k+1}^{j}\beta_i \prod_{q=j+1}^{C}\gamma_q\Big)
= \tilde\beta_k
  \sum_{j=k}^{C}\frac{\tilde\gamma_j}{\tilde\beta_j}
= \tilde\beta_k R_k.
\end{equation}
\myparagraph{Closed form for $\mathbf{W}_C$.} Combining these terms, the final weight matrix $\mathbf{W}_C$ is computed as:
\begin{equation}
\label{eq:final-closed-form-matrix-scalar}
\mathbf{W}_C
= \tilde\gamma_0 \mathbf{W}_0
\,+\, \tilde\beta_0 R_1 \mathbf{M}_0
\,+\, \sum_{t=1}^{C} \big( \eta_t \, \tilde\beta_t\, R_t\big)\, \mathbf{G}_t.
\end{equation}
\myparagraph{Closed form for $\mathbf{M}_C$.}
Since the momentum recurrence does not depend on $\mathbf{W}_t$, its closed form follows directly from \eqref{eq:m-unroll-matrix-scalar} by setting $t=C$:
\begin{equation}
\mathbf{M}_C
= \Big(\prod_{i=1}^{C}\beta_i\Big) \mathbf{M}_0
  + \sum_{t=1}^{C}\Big(\eta_t\, \mathbf{G}_t \prod_{i=t+1}^{C}\beta_i\Big).
\label{eq:m-unroll-matrix-scalar-C}
\end{equation}
Recognizing the suffix products defined in \eqref{eq:suffix-products-scalar},
\(\tilde\beta_0 = \prod_{i=1}^{C}\beta_i\) and \(\tilde\beta_t = \prod_{i=t+1}^{C}\beta_i\),
Eq.~\eqref{eq:m-unroll-matrix-scalar-C} collapses to the parallel kernel form
\begin{equation}
\label{eq:final-closed-form-momentum}
\mathbf{M}_C
= \tilde\beta_0\, \mathbf{M}_0
\,+\, \sum_{t=1}^{C} \big( \eta_t\, \tilde\beta_t \big) \, \mathbf{G}_t.
\end{equation}
Unlike $\mathbf{W}_C$, the momentum kernel involves no cumulative ratio $R_t$: each gradient $\mathbf{G}_t$ is propagated to the chunk end through a single suffix product $\tilde\beta_t$, since only one decay channel ($\beta$) acts on $\mathbf{M}_t$.

This formulation allows us to ``jump'' directly from $t=0$ to $t=C$ without materializing intermediate states. To ensure numerical stability, particularly with long chunk lengths $C$, we perform the cumulative product computations ($\tilde\beta, \tilde\gamma$) in log-space and convert back to linear space only for the final summation.

\subsection{Stability of the Cumulative Kernel \texorpdfstring{$R_t$}{Rt}}
\label{app:stability}

We analyze the magnitude of the cumulative kernel $R_t$ under two regimes that bound the practical operating
range of our input-dependent dynamics: (i)~near-unit retention
($\beta_t \to 1$ and $\gamma_t \to 1$), as encouraged by the timescale
parameterization (Eq.~\ref{eq:beta_parameterization}) for long-range memory, and
(ii)~aggressive decay ($\gamma_t$ small).

\myparagraph{Near-unit retention regime.}
When $\beta_t \to 1$ and $\gamma_t \to 1$ for all $t$, the suffix products
satisfy $\tilde{\beta}_i \to 1$ and $\tilde{\gamma}_i \to 1$, so
$\tilde{\gamma}_i / \tilde{\beta}_i \to 1$ for every $i$. Substituting into the
definition of $R_t$ gives
\begin{equation}
R_t \;\to\; \sum_{i=t}^{C} 1 \;=\; C - t + 1 \;\le\; C.
\end{equation}
The weight kernel $\mathcal{K}^W_t = \eta_t \, \tilde{\beta}_t \, R_t$ is therefore bounded by
\begin{equation}
\mathcal{K}^W_t \;\le\; \eta_t \cdot 1 \cdot C.
\end{equation}
Because $\eta_t$ is sigmoid-gated and scaled by $\eta_{\text{base}}$
(Eq.~\ref{eq:eta_parameterization}), the per-token kernel contribution is bounded by
$\eta_{\text{base}} \cdot C$. With our default
$\eta_{\text{base}} = 10^{-2}$ and $C = 512$, this yields a worst-case bound of
$5.12$. The momentum kernel $\mathcal{K}^M_t = \eta_t \, \tilde{\beta}_t$ is more tightly
bounded by $\eta_{\text{base}}$.

\myparagraph{Aggressive decay regime.}
When $\gamma_t$ is small, the suffix product $\tilde{\gamma}_i$ decays
geometrically as $i$ moves away from the chunk end, and so does the ratio
$\tilde{\gamma}_i / \tilde{\beta}_i$. The sum $R_t$ is then dominated by its
last few terms ($i$ close to $C$) and remains $\mathcal{O}(1)$.

\myparagraph{Coupled parameterization.}
The decay parameterization $\gamma_t = 1 - \eta_t \alpha_t$
(Eq.~\ref{eq:decay_derivation}) enforces a structural coupling: large updates
($\eta_t$ large) automatically increase forgetting ($\gamma_t$ smaller),
preventing carry inflation across chunks. This analysis is consistent with
the numerical verification reported in Sec.~\ref{sec:closed_form_parallelization}: over a
$128$-chunk trajectory ($C=512$, $65{,}536$ tokens), the relative $\ell_2$
deviation from the fully sequential reference is below $2 \times 10^{-6}$ on
both $\mathbf{W}_C$ and $\mathbf{M}_C$.

\subsection{Single Backward Pass for Both Aggregates}
\label{app:single-backward}

Proposition~\ref{prop:exactness} expresses both chunk-end states through a
sum over per-token gradients
$\mathbf{G}_t = -\nabla_{\mathbf{W}} \mathcal{L}_t \big|_{\mathbf{W}_0}$:
\begin{align}
\sum_{t=1}^{C} \mathcal{K}^W_t \, \mathbf{G}_t
  &\;=\; -\sum_{t=1}^{C} \mathcal{K}^W_t \, \nabla_{\mathbf{W}} \mathcal{L}_t
        \big|_{\mathbf{W}_0}, \\
\sum_{t=1}^{C} \mathcal{K}^M_t \, \mathbf{G}_t
  &\;=\; -\sum_{t=1}^{C} \mathcal{K}^M_t \, \nabla_{\mathbf{W}} \mathcal{L}_t
        \big|_{\mathbf{W}_0},
\end{align}
where the kernels $\mathcal{K}^W_t = \eta_t \, \tilde{\beta}_t \, R_t$ and
$\mathcal{K}^M_t = \eta_t \, \tilde{\beta}_t$ differ only by the scalar factor $R_t$.

\myparagraph{Chain-rule decomposition.}
Each per-token loss has the form
$\mathcal{L}_t = \mathcal{L}\!\left(f_{\mathbf{W}_0}(\boldsymbol{k}_t), \boldsymbol{v}_t\right)$.
Applying the chain rule,
\begin{equation}
\nabla_{\mathbf{W}} \mathcal{L}_t \big|_{\mathbf{W}_0}
\;=\;
-\left(\frac{\partial f_{\mathbf{W}_0}(\boldsymbol{k}_t)}
           {\partial \mathbf{W}}\right)^{\!\top} \boldsymbol{g}_t,
\qquad
\boldsymbol{g}_t \;:=\;
-\frac{\partial \mathcal{L}_t}
     {\partial f_{\mathbf{W}_0}(\boldsymbol{k}_t)}.
\end{equation}
The Jacobian
$\left(\partial f_{\mathbf{W}_0}(\boldsymbol{k}_t) / \partial \mathbf{W}\right)^{\!\top}$
depends only on $(\mathbf{W}_0, \boldsymbol{k}_t)$ and is independent of the
scalar kernels.

\myparagraph{Shared backward, scalar re-weighting.}
Both aggregates therefore reduce to scalar-weighted contractions over the
same set of activation gradients $\{\boldsymbol{g}_t\}_{t=1}^{C}$ and per-token inputs $\{\boldsymbol{k}_t\}_{t=1}^{C}$, differing only in
the scalar weight $\mathcal{K}^W_t$ versus $\mathcal{K}^M_t$ applied per token. A single
autograd backward pass through the per-token losses produces $\{\boldsymbol{g}_t\}$.
Both aggregates are then constructed by reusing these intermediates with
the appropriate scalar weighting. The marginal cost of producing
$\mathbf{M}_C$ in addition to $\mathbf{W}_C$ is therefore one extra
scalar-weighted contraction, not a second backward pass. This is consistent
with the throughput measurements reported in
Fig.~\ref{fig:training_throughput}, where \ours{} stays close to the throughput of
LaCT~\cite{zhang2025test} despite computing the exact closed-form kernel.

\subsection{Two Equivalent Views of the Chunk Dynamics}
\label{app:two-views}

Within a chunk, all gradients $\mathbf{G}_t$ in our formulation are
evaluated at the frozen $\mathbf{W}_0$ (Sec.~\ref{sec:closed_form_parallelization}).
Under this constraint, the closed-form chunk-end states from
Proposition~\ref{prop:exactness} admit two complementary readings.

\myparagraph{Optimization-style view.}
Equation~\eqref{eq:tokenwise_ttt_decay_momentum} reads as a token-wise state
recurrence with momentum factor $\beta_t$ and L$_2$ decay factor $\gamma_t$,
both modulated per token. This is the framing we adopt throughout the
paper, for consistency with the mini-batch and chunk-wise TTT
literature~\cite{sun2024learning, behrouz2024titans, zhang2025test}. This view is
accurate at the inter-chunk level, where $\mathbf{M}_C$ is carried
into the next chunk and accumulates momentum over the full sequence.

\myparagraph{Kernel-weighting view.}
Equivalently, the chunk-end states can be read as scalar-weighted
aggregations of per-token gradients,
\begin{equation*}
\mathbf{W}_C
\;=\; \tilde{\gamma}_0 \mathbf{W}_0
   + \tilde{\beta}_0 R_1 \mathbf{M}_0
   + \sum_{t=1}^{C} \mathcal{K}^W_t \, \mathbf{G}_t,
\qquad
\mathbf{M}_C
\;=\; \tilde{\beta}_0 \mathbf{M}_0
   + \sum_{t=1}^{C} \mathcal{K}^M_t \, \mathbf{G}_t,
\end{equation*}
in which $(\mathcal{K}^W_t, \mathcal{K}^M_t)$ act as learned, time-aware kernel weights over
per-token gradient contributions to the chunk-end states.

The two views are mathematically equivalent for the chunk-end pair
$(\mathbf{W}_C, \mathbf{M}_C)$. They differ only in emphasis. The
optimization-style view is the natural reading across chunk boundaries,
where momentum is carried forward. The kernel-weighting view is the
natural reading within a chunk, where the
frozen-$\mathbf{W}_0$ constraint makes ``momentum'' and ``decay'' equivalent
to a learned per-token weighting of gradient contributions.

\section{Experiment Continued}

\subsection{Training Details}
\label{sec:training_details}

\begin{wraptable}{r}{.47\linewidth}
\vspace{-4mm}
\caption{Model hyperparameters.} \vspace{-2mm}
\label{tab:hyperparameters}
\begin{center}
\resizebox{\linewidth}{!}{
\begin{tabular}{rcc}
\toprule
 & \multicolumn{2}{c}{Model} \\
  & 340M & 1.3B  \\ \midrule
Number of layers & \multicolumn{2}{c}{24}  \\
Feedforward block multiplier & \multicolumn{2}{c}{4} \\ 
Total hidden size & 1024 & 2048 \\ 
Number of heads & 8 & 16 \\
\midrule
Sequence length & 2048 & 2240 \\
Effective Batch size &  \multicolumn{2}{c}{256} \\ % [ERRATA-5]
Learning rate & \multicolumn{2}{c}{$10^{-3}$}  \\
Warmup steps & \multicolumn{2}{c}{1024} \\
Minimum learning rate ratio & \multicolumn{2}{c}{0.1} \\
Max norm clipping & \multicolumn{2}{c}{1.0} \\
Std.~of weight initializers & \multicolumn{2}{c}{0.02} \\
\midrule
TTT base learning rate $\eta_{\text{base}}$ & \multicolumn{2}{c}{0.01} \\
TTT momentum timescale $\tau$ & \multicolumn{2}{c}{32} \\
TTT decay strength $\alpha_{\text{base}}$ & \multicolumn{2}{c}{0.1} \\
\bottomrule
\end{tabular}}
\end{center}
\vspace{-5mm}
\end{wraptable}
Table \ref{tab:hyperparameters} shows the training and model hyperparameters used to train the 340M and 1.3B parameter models.
We train with an effective batch size of 256 and a sequence length of 2048 for 28,672 steps, amounting to 15B tokens. The 1.3B models are trained with a slightly increased sequence length of 2240, which increases the training token count to 16B (for simplicity, in the main text, we refer to both as trained for 15B tokens). 
We implement our expressive chunk-wise TTT in plain PyTorch without custom Triton kernels.  Training the 340M and 1.3B models takes about 132 and 348 total GPU hours on H100 GPUs, respectively. Following~\citet{zhang2025test}, we apply L2 normalization to the fast-weight matrices at each chunk boundary as a standard stabilizer.

In our video understanding experiments, we instantiate \ours (parameterized by a SwiGLU MLP) as a parallel
branch fused with the existing Qwen3VL self-attention layers. \ours employs its own query, key, and value projection layers. Following standard practice~\cite{alayrac2022flamingo,dalal2025one}, we gate the output of \ours with a learned vector $\varepsilon \in \mathbb{R}^d$ with tanh activation. All entries of $\varepsilon$ are initialized to 0.1, such that $\tanh(\varepsilon)$ is close to 0 at the beginning of fine-tuning. We train only the TTT parameters on a 43K-sample subset of LLaVA-Video-178K~\cite{zhang2024video}.

\subsection{Evaluation Details}
\label{sec:evaluation_details}

\myparagraph{Commonsense Reasoning.} Following prior work~\cite{gu2024mamba,yang2024gated,irie2025blending}, we evaluate our model on multiple commonsense reasoning benchmarks: PIQA~\cite{bisk2020piqa}, HellaSwag \citep[Hella.;][]{zellers2019hellaswag}, WinoGrande \citep[Wino.;][]{sakaguchi2020winogrande}, ARC-easy (ARC-e) and ARC-challenge (ARC-c) \citep{clark2018think}, Wikitext \citep[Wiki.;][]{merity2016pointer}, and LAMBADA \citep[LMB.;][]{paperno2016lambada}. All evaluations are performed using \texttt{lm-evaluation-harness} \citep{gao2021framework}.

\myparagraph{In-context Retrieval.} Following prior work~\citep{yang2024parallelizing,arora2023language,irie2025blending},
we focus on three tasks: FDA \citep{arora2023language}, SWDE \citep{lockard2019openceres}, and SQuAD \citep{rajpurkar2018know}. For SQuAD and SWDE, we use \texttt{lm-evaluation-harness} \cite{gao2021framework} for evaluation. For FDA, we follow \citet{irie2025blending} and use the evaluation script from \citet{arora2024just}.

\myparagraph{Length Extrapolation.} 
We first assess whether the models can maintain stable language modeling performance when processing sequences longer than their training context (2K tokens). Following~\citet{yang2024gated}, we conduct experiments on six diverse long-context benchmarks: PG19~\cite{rae2019compressive}, GovReport~\cite{huang2021efficient}, QMSum~\cite{zhong2021qmsum}, NarrativeQA~\cite{kovcisky2018narrativeqa}, Qasper~\cite{dasigi2021dataset}, and CodeParrot. To test effective capacity, we utilize the S-NIAH-1 (passkey retrieval) 
and S-NIAH-2 (numerical needle in haystack) tasks from RULER's 
Needle-In-A-Haystack Single (NIAH-S) benchmark suite 
\citep{hsieh2024ruler}. 
Finally, we evaluate on 14 real-world long-context tasks from Longbench \citep{bai2024longbench}, including: narrative comprehension (Narrative QA \citep{kovcisky2018narrativeqa}), scientific understanding (QasperQA \citep{dasigi2021dataset}), multi-hop reasoning (MultiField QA, HotpotQA \citep{yang2018hotpotqa}, 2WikiMulti QA \citep{ho2020constructing}, Musique \citep{trivedi2022musique}), document summarization (GovReport \citep{huang2021efficient}, QMSum \citep{zhong2021qmsum}, MultiNews \citep{fabbri2019multi}), and various specialized tasks (TRec \citep{li2002learning}, Trivia QA \citep{joshi2017triviaqa}, SamSum \citep{gliwa2019samsum}, LCC \citep{guo2023longcoder}, and RepoBench-P \citep{liu2023repobench}). All evaluations, except for the stability analysis, are performed using \texttt{lm-evaluation-harness} \citep{gao2021framework}.

\myparagraph{Video Understanding.} We evaluate on two standard video understanding benchmarks: VideoMMMU~\cite{hu2025video} and LongVideoBench~\cite{wu2024longvideobench}. Following the Qwen3VL~\cite{Qwen3-VL} protocol, we employ a uniform sampling strategy to extract a fixed sequence of frames from each video. We configure the frame count and token budget to balance temporal range with spatial resolution: for VideoMMMU, we sample 512 frames with a maximum of 256 tokens per frame, and for LongVideoBench, 256 frames with 512 tokens.

\myparagraph{Hybrid Baselines.} We benchmark against two competitive hybrid architectures: HQLT~\cite{irie2025blending} and LaCT~\cite{zhang2025test}. For HQLT, which combines DeltaNet~\cite{schlag2021linear} with window attention, we select the \textit{synchronous} variant where both components process inputs simultaneously, which the authors report as the best-performing setting. For LaCT, which fuses chunk-wise TTT (parameterized by a SwiGLU MLP) with window attention, we evaluate the variant using Muon optimization for the TTT update rule, consistent with the best-performing configuration provided in the official implementation.

\subsection{Ablation Study}
\label{sec:ablation}

We validate the design of \ours along four axes. 
\textbf{(i)}~The proposed closed-form kernel is the source of accuracy gains over
chunk-wise baselines that average per-token factors, and finer chunk granularity
narrows the residual gap to the ideal token-wise update
(Table~\ref{tab:abl_kernel}).
\textbf{(ii)}~Both branches of the hybrid architecture and the post-chunk
normalization are individually necessary, as removing any one of them collapses
either retrieval or long-context performance (Table~\ref{tab:abl_components}).
\textbf{(iii)}~The token-wise momentum and weight-decay terms recovered exactly
by our kernel reduce perplexity over a no-momentum baseline
(Table~\ref{tab:abl_inner}).
\textbf{(iv)}~The method is robust to the choice of base learning rate and base
decay over a wide range (Table~\ref{tab:abl_sensitivity}).

\begin{table}[h!]
\centering
\small
\caption{Effect of the closed-form kernel and chunk granularity (340M parameters,
15B tokens). The averaged-factor baseline collapses per-token
$\beta_t,\gamma_t$ into chunk-level scalars, whereas our kernel preserves the full
per-token dynamics in closed form.}
\label{tab:abl_kernel}
\begin{tabular}{lccc}
\toprule
Configuration & Wiki ppl $\downarrow$ & LMB ppl $\downarrow$ & Avg $\uparrow$ \\
\midrule
Averaged-factor baseline ($C{=}512$) & 26.9 & 25.9 & 48.6 \\
Closed-form kernel, $C{=}1024$       & 25.6 & 26.2 & 48.4 \\
Closed-form kernel, $C{=}512$        & \textbf{25.5} & \textbf{25.0} & \textbf{49.1} \\
\bottomrule
\end{tabular}
\end{table}

\myparagraph{Effect of the closed-form kernel and chunk granularity (Table~\ref{tab:abl_kernel}).}
We compare against a chunk-wise baseline that averages the per-token decay and
momentum factors into chunk-level scalars, following the simplification used in
LaCT~\citep{zhang2025test}. Replacing the averaged factors with our closed-form
kernel reduces Wiki perplexity from 26.9 to 25.5 and improves average accuracy
from 48.6\% to 49.1\% at $C{=}512$. We then sweep the chunk size: $C{=}1024$
already outperforms the averaged baseline (Wiki ppl 25.6, Avg 48.4\%), and
$C{=}512$ further improves both metrics. This trend matches the theoretical
prediction that $C \to 1$ recovers the token-wise update exactly. Smaller chunks
narrow the gap but reduce hardware utilization. We therefore use $C{=}512$ in the
main experiments as a trade-off between efficiency and quality.

\begin{table}[h!]
\centering
\small
\caption{Component ablation at the 1.3B scale on S-NIAH-1 (accuracy $\uparrow$
at three context lengths) and LongBench (average score $\uparrow$). The full
model is trained for 15B tokens at 2K context. All ablations use the same
training data and hyperparameters.}
\label{tab:abl_components}
\begin{tabular}{lcccc}
\toprule
Configuration & S-NIAH-1 2K & S-NIAH-1 8K & S-NIAH-1 16K & LongBench \\
\midrule
SWA only (no TTT branch)         & 26.8 &  6.8 &  2.8 & 11.2 \\
TTT only (no SWA branch)         &  0.0 &  0.0 &  0.0 &  1.5 \\
Full, w/o post-chunk norm        & \textbf{100.0} & \textbf{99.8} & \textbf{98.8} & 13.4 \\
Full, w/o fusion gate            & 99.6 & 88.8 & 86.6 & 13.3 \\
\textbf{Full \textsc{E$^2$-TTT$_{\text{SwiGLU}}$}} & 95.6 & 95.4 & 93.6 & \textbf{14.1} \\
\bottomrule
\end{tabular}
\end{table}

\myparagraph{Branch contributions (Table~\ref{tab:abl_components}).}
We isolate the role of each component of the hybrid architecture at the 1.3B
scale, evaluating on Single-NIAH-1 at three context lengths and on LongBench.
Sliding-window attention alone reaches only
6.8\% on S-NIAH-1 at 8K tokens, confirming that local attention cannot solve
long-range retrieval. The TTT branch alone (no SWA) collapses to 0.0\%
across all S-NIAH lengths, because the chunk-wise output step cannot resolve
within-chunk dependencies. This is the ``blind spot'' that our hybrid design
addresses. With both branches, S-NIAH-1 reaches 95.4\% at 8K and
93.6\% at 16K. Post-chunk normalization trades a small amount of synthetic-retrieval
accuracy ($99.8$ versus $95.4$ on S-NIAH-1 at 8K) for real-world long-context
performance (LongBench $14.1$ versus $13.4$). Removing the fusion gate
costs accuracy beyond 2K on both axes.

\begin{table}[h!]
\centering
\small
\caption{Effect of the inner-loop optimizer terms recovered by our kernel
(340M, $C{=}512$, 15B tokens). The baseline retains only the input-dependent
learning rate $\eta_t$.}
\label{tab:abl_inner}
\begin{tabular}{lccc}
\toprule
Inner-loop update rule & Wiki ppl $\downarrow$ & LMB ppl $\downarrow$ & Avg $\uparrow$ \\
\midrule
$\eta_t$ only (no momentum, no decay) & 26.6 & 26.4 & 48.7 \\
$\eta_t$ + momentum + weight decay    & \textbf{25.5} & \textbf{25.0} & \textbf{49.1} \\
\bottomrule
\end{tabular}
\end{table}

\myparagraph{Inner-loop momentum and weight decay (Table~\ref{tab:abl_inner}).} We test whether the input-dependent
momentum and decay terms contribute, by training
two variants with $C{=}512$: a no-momentum/no-decay baseline (which retains
only the input-dependent learning rate) and the full
update rule. Adding momentum and weight decay reduces perplexity
on both Wiki and LMB.

\begin{table}[h!]
\centering
\small
\caption{Sensitivity of \textsc{E$^2$-TTT$_{\text{MLP}}$} (340M) to the base
hyperparameters of the input-dependent dynamics.}
\label{tab:abl_sensitivity}
\begin{tabular}{lccc}
\toprule
Configuration & Wiki ppl $\downarrow$ & LMB ppl $\downarrow$ & Avg $\uparrow$ \\
\midrule
\multicolumn{4}{l}{\emph{Base learning rate $\eta_{\text{base}}$ ($\alpha_{\text{base}}{=}10^{-1}$)}} \\
\quad $\eta_{\text{base}} = 10^{-1}$ & 28.8 & 30.3 & 46.5 \\
\quad $\eta_{\text{base}} = 10^{-2}$ \emph{(default)} & 25.5 & 25.0 & 49.1 \\
\quad $\eta_{\text{base}} = 10^{-3}$ & 26.1 & 23.0 & 49.1 \\
\midrule
\multicolumn{4}{l}{\emph{Base decay $\alpha_{\text{base}}$ ($\eta_{\text{base}}{=}10^{-2}$)}} \\
\quad $\alpha_{\text{base}} = 10^{-1}$ \emph{(default)} & 25.5 & 25.0 & 49.1 \\
\quad $\alpha_{\text{base}} = 10^{-2}$ & 25.8 & 23.6 & 49.0 \\
\quad $\alpha_{\text{base}} = 10^{-3}$ & 25.7 & 22.9 & 49.0 \\
\bottomrule
\end{tabular}
\end{table}

\myparagraph{Sensitivity to base hyperparameters (Table~\ref{tab:abl_sensitivity}).}
The input-dependent dynamics expose a base learning rate $\eta_{\text{base}}$
and a base decay $\alpha_{\text{base}}$ that scale the per-token sigmoid
gates. We sweep each over two orders of magnitude. Performance is essentially
invariant to $\alpha_{\text{base}}$ in the tested range and stable for
$\eta_{\text{base}} \in \{10^{-3}, 10^{-2}\}$. Only the aggressive
$\eta_{\text{base}}{=}10^{-1}$ degrades training, consistent with classical
optimization where overly large step sizes destabilize updates. The defaults
$\eta_{\text{base}}{=}10^{-2}$ and $\alpha_{\text{base}}{=}10^{-1}$ used
throughout the paper are not finely tuned.

\subsection{Analysis of the Learned Coefficients}
\label{sec:coeff_analysis}

The ablations of Appendix~\ref{sec:ablation} remove a term from the update rule
and retrain. A complementary question is whether the three linear heads of
Eqs.~\eqref{eq:eta_parameterization}--\eqref{eq:decay_derivation} actually use
their input-dependence once trained, or whether they instead settle near
constants or silently compensate for one another. We answer this with two
measurements that require no retraining: both run post-hoc on the single
trained 1.3B \textsc{E$^2$-TTT$_{\text{SwiGLU}}$} checkpoint used
throughout the main experiments.

\myparagraph{The coefficients are not near-constant.} We measure dispersion on
the three positive quantities whose zero value means ``no effect'': the step
size $\eta_t$, the forgetting rate $-\log \beta_t$, and the decay rate
$1 - \gamma_t = \eta_t \alpha_t$. Over WikiText-2, the median coefficient
of variation across all (layer, head) pairs is $0.95$ for $\eta_t$, $0.46$ for
$-\log \beta_t$, and $1.29$ for $1 - \gamma_t$. Fewer than $1\%$ of pairs fall
below $0.1$. Compounded over a chunk of $C{=}512$ tokens, the spread is substantial:
the fast-weight carry $\prod_t \gamma_t$ spans $0.73$--$1.00$
and the momentum carry $\prod_t \beta_t$ spans a factor of $129$ between its
$5$th and $95$th percentiles. This is the per-token variation that a
chunk-averaged update rule discards.

\myparagraph{Causal test: shuffling the coefficients along the token axis.} A
wide distribution alone does not establish that the values are aligned
with the tokens that produced them. For each head we therefore permute its
scalar along the token axis, independently per sequence and per head. The
permutation preserves the marginal distribution exactly and destroys only the
correspondence between a coefficient and its token. Any degradation can
therefore be attributed to input-dependence alone. For $\alpha$ we permute the head output
and recompute $\gamma_t = 1 - \eta_t \alpha_t$, which leaves $\eta_t$ aligned.
All variants are scored with the same \texttt{lm-evaluation-harness} invocation
used for Table~\ref{tab:commonsense_main}.

\begin{table}[h!]
\centering
\small
\caption{Shuffling each coefficient head along the token axis on the trained
1.3B \textsc{E$^2$-TTT$_{\text{SwiGLU}}$} checkpoint. The
permutation leaves the marginal distribution of each scalar unchanged and
removes only its alignment with the input.}
\label{tab:abl_shuffle}
\begin{tabular}{lccc}
\toprule
Shuffled head & Wiki ppl $\downarrow$ & $\Delta$ ppl & S-NIAH-1 16K $\uparrow$ \\
\midrule
--- (baseline)   & \textbf{19.94} & ---   & \textbf{93.6} \\
$\eta$           & 20.39 & $+0.45$ &  6.0 \\
$\beta$          & 20.05 & $+0.11$ & 86.2 \\
$\alpha$         & 19.99 & $+0.05$ & 91.8 \\
All three        & 20.54 & $+0.60$ &  4.0 \\
\bottomrule
\end{tabular}
\end{table}

Every shuffle degrades performance, and $\eta$ has by far the largest effect.
Shuffling $\eta$ alone costs $+0.45$ perplexity and collapses retrieval ($93.6 \to 6.0$ on
S-NIAH-1 at 16K), confirming that the step size carries most of the
write-selectivity of the inner loop. The joint intervention costs approximately
the sum of the individual ones ($+0.60$ versus $0.45 + 0.11 + 0.05$),
indicating that the three heads contribute independently. If the heads
substituted for one another, shuffling one alone would be cheap because the
other two would absorb its role, and the joint cost would exceed the sum.
This additivity can only be assessed on perplexity, since the $\eta$ shuffle
alone already drives S-NIAH accuracy to its floor.

The $\beta$ and $\alpha$ shuffles have smaller effects than the $\eta$ shuffle
partly by construction. Because Eq.~\eqref{eq:decay_derivation} couples the decay
to the step size, permuting only the $\alpha$ head leaves the $\eta$-driven
component of $\gamma_t$ aligned. Nonetheless, Table~\ref{tab:abl_inner} independently shows that the momentum
and decay terms matter, as removing them from the update rule increases Wiki
perplexity from 25.5 to 26.6 at 340M. Taken together, the two measurements
show that the per-token scalars are input-dependent and non-redundant, which
supports aggregating them exactly with the closed-form kernel
(Sec.~\ref{sec:closed_form_parallelization}) instead of averaging them per chunk.

\subsection{Throughput and Inference Cost}
\label{sec:throughput}

\begin{figure}[h!]
    \centering
    %\vspace{-6mm}
    \includegraphics[width=.6\linewidth]{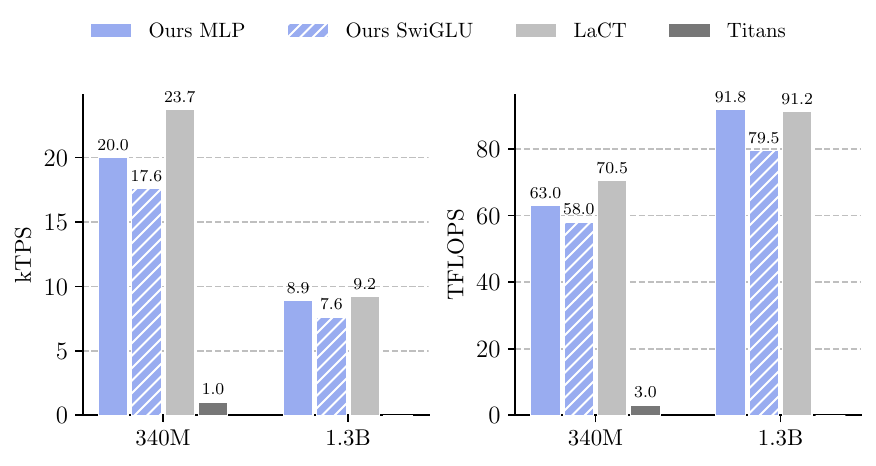}\vspace{-2pt}
    \caption{Training throughput on a single H100 GPU at the 340M and 1.3B scales.}
    \label{fig:training_throughput}
    \vspace{-2mm}
\end{figure}

To validate the hardware efficiency of our proposed method, we evaluate training throughput (measured in kilo-tokens per second, kTPS) and utilization (TFLOPS) on a single H100 64GB GPU. We test two model scales (340M and 1.3B parameters) on 2K sequences, using batch sizes of 4 and 2, respectively. We compare \ours against Titans~\citep{behrouz2024titans} and LaCT~\citep{zhang2025test}. 
All baselines are implemented in PyTorch under
identical training conditions. As no official Titans implementation was
available at the time of writing, we report a self-reproduction following the
architecture described in \citet{behrouz2024titans}. We implemented both the
scan-based formulation and the memory-optimized dual form~\citep{sun2024learning}
and report the latter, which we found slightly faster in PyTorch. 
As shown in Figure~\ref{fig:training_throughput}, Titans is severely
throughput-bound: at 340M it achieves less than 2~kTPS, and at 1.3B it goes
out-of-memory. This is consistent with the structure of the dual form for
nonlinear inner models such as TTT-MLP: each inner-model layer is
reformulated as a masked attention-like computation
(Appendix~A.3, \citet{sun2024learning}), and this machinery lies on the
outer-loop autograd path, inflating both activation memory and backward-pass
cost. The dual form does eliminate per-token state materialization in principle,
but for TTT-MLP the small inner mini-batch size (64 in our setting) combined
with these masked-matmul overheads dominates throughput. These costs are absent
under chunk-wise output (Eq.~\ref{eq:chunkwise_output}), where the per-chunk forward and backward
operate on a single fast-weight state.
In contrast, both \ours variants stay close to the throughput of the highly optimized
LaCT baseline despite computing the exact closed-form kernel rather than a
single averaged factor. Our kernel adds log-space suffix products, ratio sums,
and a scalar re-weighting that yields $\mathbf{M}_C$ from the same backward pass
that produces $\mathbf{W}_C$. This bookkeeping is cheap relative to the
fast-weight forward and backward passes.
Relative to LaCT, the exact kernel costs $15.6\%$ throughput at 340M and
only $3.3\%$ at 1.3B for \oursmlp. The
overhead shrinks with scale because the $d^2$ fast-weight forward and backward
passes increasingly dominate the scalar bookkeeping. 
Exact per-token dynamics therefore come at a small cost in
training throughput.

\begin{table}[h!]
\centering
\small
\caption{Inference cost at 1.3B on a single H100 GPU, at batch sizes
$1$ and $8$. Decode latency is the mean per-token time over $512$ steps after
warm-up (median of three runs). All recurrent models are flat in both latency
and memory between 2K and 16K, while attention is not.}
\label{tab:inference_cost}
\begin{tabular}{lcccc}
\toprule
& \multicolumn{2}{c}{Decode (ms/token) $\downarrow$} & \multicolumn{2}{c}{Peak mem.\ (GB) $\downarrow$} \\
\cmidrule(lr){2-3} \cmidrule(lr){4-5}
Model & 2K & 16K & 2K & 16K \\
\midrule
\multicolumn{5}{l}{\textit{Batch size 1}} \\
Transformer++   & 14.8 & 15.0 & 3.25 & 6.10 \\
Titans          & 56.1 & 56.2 & 3.41 & 3.51 \\
LaCT            & 36.8 & 36.5 & 3.33 & 3.42 \\
\oursmlp        & 33.8 & 33.6 & 3.43 & 3.52 \\
\oursglu        & 34.0 & 33.9 & 3.62 & 3.71 \\
\midrule
\multicolumn{5}{l}{\textit{Batch size 8}} \\
Transformer++   & 14.8 & 59.9 & 7.50 & 29.90 \\
Titans          & 56.5 & 56.6 & 6.40 & 6.40 \\
LaCT            & 38.2 & 38.8 & 7.10 & 7.10 \\
\oursmlp        & 33.7 & 33.8 & 6.86 & 6.86 \\
\oursglu        & 34.3 & 34.4 & 8.08 & 8.08 \\
\bottomrule
\end{tabular}
\end{table}

\myparagraph{Inference cost.} 
Table~\ref{tab:inference_cost} reports decode latency and peak memory at 1.3B.
Attention is faster at short sequences thanks to its optimized kernels, but it is
the only model whose cost grows with length ($14.8 \to 59.9$~ms/token and
$7.50 \to 29.90$~GB between 2K and 16K at batch $8$), while all recurrent models
are flat in both because their state is of constant size. Among TTT methods,
Titans is the slowest to decode, for the same reason it is slow to train. Both
\ours variants decode slightly faster than LaCT at comparable memory, as the
per-chunk scalar bookkeeping is amortized over $C$ tokens and adds negligible
cost to the fast-weight forward pass. The one exception is the peak memory of \oursglu at batch $8$ ($8.08$
against LaCT's $7.10$~GB), which comes from its three-matrix fast-weight network 
and the extra re-weighted aggregation (Algorithm~\ref{alg:e2ttt}). 
\oursmlp, whose fast-weight network is smaller, stays below LaCT at batch $8$
($6.86$ versus $7.10$~GB) and within $0.1$~GB of it at batch $1$.

\subsection{Detailed S-NIAH Results}
\label{app:s-niah-full}

Section~\ref{sec:exp_extrapolation} of the main paper reports S-NIAH-1
and S-NIAH-2 results up to 16K for the four methods shown in
Figure~\ref{fig:single_niddle}. Here we provide the corresponding
numerical results, including two additional baselines not covered in
the main figure: Mamba2~\cite{mamba2} and E2E-TTT~\cite{e2e_ttt}.

\begin{table*}[h!]
    \centering
    \small
    \caption{Zero-shot performance on the RULER S-NIAH benchmark suite 
    for 1.3B models. Training context is 2K tokens; columns to the right
    of 2K are extrapolation. \oursglu retains a substantial margin over
    all baselines at every extrapolation length. Best
    result per column in bold.}
    \setlength\tabcolsep{4pt}
    %\resizebox{\linewidth}{!}{
    \begin{tabular}{l|cccccc|cccccc}
       \toprule
       Method
       & \multicolumn{6}{c}{S-NIAH-1}
       & \multicolumn{6}{c}{S-NIAH-2} \\
       & 512 & 1K & 2K & 4K & 8K & 16K
       & 512 & 1K & 2K & 4K & 8K & 16K \\
       \midrule
       Mamba2
       & \textbf{100.0} & \textbf{100.0} & 98.6 & 63.8 & 31.6 & 13.0
       & \textbf{100.0} & \textbf{99.8} & \textbf{99.4} & 45.2 & 27.4 & 5.0 \\

       HQLT
       & \textbf{100.0} & 90.0 & 72.6 & 50.2 & 37.4 & 25.2
       & \textbf{100.0} & 80.0 & 57.0 & 37.2 & 15.4 & 4.4 \\

       LaCT
       & \textbf{100.0} & 99.4 & \textbf{100.0} & 81.0 & 37.4 & 3.0
       & \textbf{100.0} & 96.8 & 91.4 & 80.6 & 5.6 & 0.0 \\

       E2E-TTT
       & \textbf{100.0} & 60.0 & 27.6 & 12.6 & 6.4 & 3.0
       & \textbf{100.0} & 68.4 & 29.8 & 17.4 & 10.2 & 3.4 \\

       \oursglu
       & \textbf{100.0} & 96.8 & 95.6 & \textbf{95.0} & \textbf{95.4} & \textbf{93.6}
       & \textbf{100.0} & 99.0 & 96.8 & \textbf{87.6} & \textbf{40.6} & \textbf{6.0} \\

       \oursmlp
       & \textbf{100.0} & 95.6 & 93.2 & 89.2 & 88.2 & 85.0
       & \textbf{100.0} & 94.2 & 79.6 & 32.6 & 9.8 & 4.0 \\
       \bottomrule
    \end{tabular}%}
\end{table*}

E2E-TTT~\cite{e2e_ttt} is a recent chunk-wise TTT method that adopts a
different inner-loop objective from the key--value binding formulation
used in our work and in LaCT~\cite{zhang2025test}. The original paper
reports results only within the training context length and provides a
JAX implementation. We re-implemented E2E-TTT in PyTorch following the
architecture and update rule of the original paper, and trained it
under our identical data, tokenizer, and optimization setup. As a sanity check, our
re-implementation reaches ceiling accuracy with a sequence length of 512, consistent with other models.
\ours{} retains a substantial margin over all baselines at every
extrapolation length on both S-NIAH variants. At 16K on S-NIAH-1,
\oursglu{} reaches $93.6\%$ while all baselines fall below $26\%$.
On S-NIAH-2, \oursglu{} remains the leading method through 16K, whereas
all sub-quadratic baselines drop below 30\% from 8K onward.

\subsection{Training at a Larger Budget}
\label{app:scaling}

To verify training feasibility at a larger budget, we additionally trained \oursglu at 1.3B parameters
for 100B tokens ($6.7\times$ the budget of every other run in this paper) at a
4K training context, leaving the fast-weight network, the sliding-window size
(512), the chunk size (512), the data source and the tokenizer unchanged. 
Table~\ref{tab:scaling_tokens} reports the resulting checkpoint. Perplexity and
commonsense accuracy improve substantially over the 15B-token model, and
in-context retrieval rises from $43.6\%$ to $58.1\%$. 
This run serves only as a feasibility check. 
Since the two rows differ in both the token budget and the training context, the gains cannot be attributed to the larger budget alone. 
It does, however, show that \ours trains stably at a budget $6.7\times$ larger than that of the main experiments.

\begin{table}[h!]
\centering
\small
\caption{A 1.3B \oursglu{} checkpoint trained for 100B tokens at a 4K
training context, shown next to the 15B-token checkpoint used throughout the main
experiments. The Commonsense and
Retrieval columns average the task sets of
Tables~\ref{tab:commonsense_main} and~\ref{tab:incontext_retrieval_main};
retrieval inputs are truncated to 2K and therefore fit inside the training
context of both rows.}
\label{tab:scaling_tokens}
\setlength\tabcolsep{6pt}
\begin{tabular}{cc|cc|cc}
\toprule
Training & Train & \multicolumn{2}{c|}{Perplexity $\downarrow$}
& Commonsense & Retrieval \\
tokens & ctx. & Wiki. & LMB. & Avg.\ $\uparrow$ & Avg.\ $\uparrow$ \\
\midrule
15B  & 2K & 19.9 & 15.8 & 53.6 & 43.6 \\
100B & 4K & 16.0 & 10.0 & 60.1 & 58.1 \\
\bottomrule
\end{tabular}
\end{table}

\section*{Impact Statement and Limitations}
\ours advances the efficiency of long-context sequence modeling by
parallelizing token-wise Test-Time Training updates. As a foundational
architecture for general-purpose language and multimodal models, it inherits
the broad set of societal considerations associated with such models, but does
not introduce risks specific to the proposed mechanism beyond those already
present in existing TTT and linear-attention variants.

We highlight two limitations of the current study. First, the
chunk-wise output step (Eq.~\ref{eq:chunkwise_output}) applies a fixed
fast-weight state to all tokens within a chunk, producing a within-chunk
``blind spot'' for fine-grained causal dependencies. We mitigate this
in practice through the sliding-window attention branch, and developing
an efficient token-wise output step remains an interesting direction
for future work. Second, following the standard evaluation protocol of
recent TTT and linear-attention work~\citep{sun2024learning,irie2025blending},
our experiments are conducted at scales up to 1.3B parameters. Extending
\ours{} to larger model sizes is a natural direction for future work.

%\newpage
%\input{checklist}

\end{document}